\documentclass[10pt,twocolumn,letterpaper]{article}

\usepackage[pagenumbers]{cvpr} 

\definecolor{cvprblue}{rgb}{0.21,0.49,0.74}
\usepackage[pagebackref,breaklinks,colorlinks,allcolors=cvprblue]{hyperref}

\def\paperID{***} 
\def\confName{CVPR}
\def\confYear{2026}

\usepackage{hyperref}
\usepackage{url}
\usepackage{booktabs}

\usepackage{colortbl} 
\usepackage{xcolor}   
\usepackage{graphicx}
\usepackage{multirow}
\usepackage{tikz}
\usepackage{wrapfig}
\usepackage{pgfplots}
\usetikzlibrary {matrix}
\usetikzlibrary{patterns}
\usepackage[dvipsnames]{xcolor}
\usepackage{subcaption}
\usepackage[T1]{fontenc}

\definecolor{lightgray}{gray}{0.95}
\definecolor{mediumgray}{gray}{0.85}
\definecolor{darkgray}{gray}{0.75}

\title{Rethinking Concept Bottleneck Models: 
From Pitfalls to Solutions }

\author{
Merve Tapli\textsuperscript{1} \quad
Quentin Bouniot\textsuperscript{2,3,4,5} \quad
Wolfgang Stammer\textsuperscript{6,7} \quad
Zeynep Akata\textsuperscript{2,3,4,5} \quad
Emre Akbas\textsuperscript{1,2,8} \\[0.5em]
\textsuperscript{1}Middle East Technical University (METU)\quad
\textsuperscript{2}Helmholtz Munich \\
\textsuperscript{3}Munich Center for Machine Learning \quad
\textsuperscript{4}Munich Data Science Institute\\
\textsuperscript{5}Technical University of Munich (TUM)\quad
\textsuperscript{6}Max Planck Institute for Informatics, SIC \\
\textsuperscript{7}RTG Neuroexplicit Models\quad
\textsuperscript{8}Robotics \& AI Center (ROMER), METU \\[0.5em]
}

\begin{document}
\maketitle
\begin{abstract}
Concept Bottleneck Models (CBMs) ground predictions in human-understandable concepts but face fundamental limitations: the absence of a metric to pre-evaluate concept relevance, the ``linearity problem'' causing recent CBMs to bypass the concept bottleneck entirely, an accuracy gap compared to opaque models, and finally the lack of systematic study on the impact of different visual backbones and VLMs. We introduce CBM-Suite, a methodological framework to systematically addresses these challenges. First, we propose an entropy-based metric to quantify the intrinsic suitability of a concept set for a given dataset. Second, we resolve the linearity problem by inserting a non-linear layer between concept activations and the classifier, which ensures that model accuracy faithfully reflects concept relevance. Third, we narrow the accuracy gap by leveraging a distillation loss guided by a linear teacher probe. Finally, we provide comprehensive analyses on how different vision encoders, vision-language models, and concept sets interact to influence accuracy and interpretability in CBMs. Extensive evaluations show that CBM-Suite yields more accurate models and provides insights for improving concept-based interpretability.
\begingroup
\renewcommand\thefootnote{}
\footnotetext{Accepted to CVPR 2026}
\footnotetext{Code is available at https://github.com/mervetapli/CBMSuite}
\endgroup
\end{abstract}    
\section{Introduction}
\label{sec:intro}

\begin{figure}[t]
    \vspace{-.5em}
    \centering
    \includegraphics[width=0.98\linewidth]{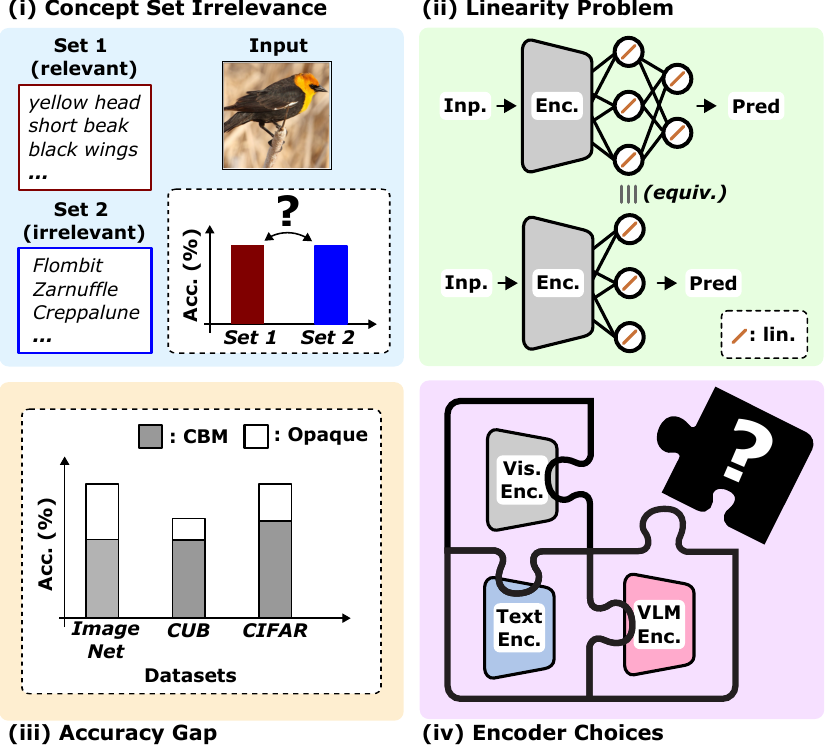}
    \caption{Common issues among CBM approaches. (i) Models can achieve high accuracy even with irrelevant or random concepts due to concept leakage. (ii) Overlooking the importance of non-linearity reduces CBMs to mere linear probes. (iii) Introducing a concept bottleneck layer often leads to an accuracy gap compared to opaque models. (iv) Although CBMs are modular by design, evaluations do not consider different encoder choices.}
    \label{fig:motivation}
\end{figure}

A long-standing challenge in vision-based machine learning is to convert powerful but opaque vision encoders into models whose predictions are interpretable and open to human intervention. Concept Bottleneck Models (CBMs) \cite{koh2020cbm, lfcbm, labo, post_hoc_cbm,srivastava2024vlgcbm} offer a natural pathway toward this goal: by projecting image representations onto a set of human-understandable concepts, followed by classification in concept space, CBMs can combine strong predictive performance with transparent reasoning. In particular, recent Vision Language Model (VLM) based approaches that harness pretrained vision-language foundation models \cite{lfcbm, post_hoc_cbm, srivastava2024vlgcbm} have been proposed as practical, high-performing solutions that eliminate the need for costly concept annotations.

However, developing trustworthy CBMs requires addressing several critical issues. The first issue is \textbf{evaluating the relevance of the concept set} (\cf \autoref{fig:motivation} (i)): CBMs may achieve high accuracy even when the chosen concept set is largely irrelevant to the dataset, as such concepts may  still encode latent correlations with class labels. 
Prior studies have documented this phenomenon, often referred to as concept or information leakage \cite{mahinpei2021leakage,margeloiu2021leakage, parisini2025leakage, makonnen2025measuring}, showing that randomly chosen or externally derived concepts may inadvertently align with image embeddings. As a result, models can produce high predictive accuracy without offering meaningful explanations, undermining the central motivation of CBMs since accuracy alone cannot guarantee interpretability.
To ensure the possibility of meaningful explanations, concept relevance should be assessed \emph{before} training, whereas current practice typically performs this evaluation only \emph{after} training through post-hoc analysis. 

The second issue is the \textbf{linearity problem} (\cf \autoref{fig:motivation} (ii)). Many recent CBM frameworks employ a purely linear mapping from feature encodings to concept activations \cite{lfcbm,post_hoc_cbm,srivastava2024vlgcbm}. In this configuration, the classifier can behave as nothing more than a linear probe on backbone embeddings, effectively ignoring the intermediate concept space. Importantly, this limitation is not dataset-specific but a structural weakness that can be observed across many existing CBM variants, including influential recent ones (\eg Label-Free CBM (LFCBM) \cite{lfcbm}). 

Next, there is the \textbf{accuracy gap} between CBMs and standard linear probes trained directly on the same vision encoder (\cf \autoref{fig:motivation} (iii)). Since CBMs restrict predictions to pass through an intermediate concept layer, they often sacrifice predictive power relative to an unconstrained probe, 
highlighting a systematic trade-off between interpretability and accuracy. In practice, this discourages practitioners from adopting CBMs in real-world settings where accuracy is critical, leaving the field with methods that are interpretable in principle but rarely competitive in practice.

Finally, there is the issue of backbone \textbf{encoder selection} and the related problem of \textbf{VLM selection} (cf. \autoref{fig:motivation} (iv)). While CBMs based on pretrained VLMs are theoretically agnostic to the choice of both the visual encoder and the underlying VLM, existing work predominantly explores only a limited subset of combinations, most commonly CLIP-based models~\citep{lfcbm, dncbm, Debole25}, without systematic justification or comparative analysis.
This narrow focus provides little guidance on how different encoders or VLMs affect accuracy and interpretability.
Because encoders and VLMs jointly determine the alingment between visual features and concept embeddings, the \emph{optimal encoder-VLM pairing} remains an open question.
Lack of thorough empirical comparison across diverse encoder models and VLMs means that the field has yet to establish best practices for model selection, potentially overlooking possible improvements.

To address the four challenges explained above, we propose CBM-Suite, a methodological framework that systematically improves the reliability and interpretability of CBMs. 
(1) For the issue of concept set relevance, we introduce an entropy-based measure to quantify the goodness of a concept set with respect to the dataset. Unlike prior approaches that only expose irrelevance after training, our method evaluates concept sets before CBM training by converting activations into probability distributions and measuring their entropy. This reveals whether concepts provide sparse, discriminative explanations or merely noisy, uninformative signals; enabling principled concept set selection and avoiding wasted computation on ineffective sets.
(2) To resolve the linearity problem, we modify the CBM architecture to ensure predictions genuinely depend on concepts. We insert a ReLU activation between two linear layers in the concept encoder, introducing the necessary non-linearity. This simple change ensures the influence of the concept set on classification accuracy. 
(3) To close the accuracy gap between CBMs and linear probes, we introduce a distillation loss that transfers knowledge from the probe into the CBM. This loss allows the CBM to inherit strong discriminative ability while retaining interpretability, narrowing the performance gap and positioning CBMs as competitive alternatives to standard classifiers in settings where both accuracy and transparency are essential.
(4) Finally, we conduct a systematic evaluation across a wide range of vision encoders (\eg, DINOv2~\cite{caron2021dinov2}, Perception Encoder~\cite{bolya2025pe}, Franca~\cite{venkataramanan2025franca}, ResNet50~\cite{he2016resnet}) and vision–language models (\eg, CLIP~\cite{clip}, SigLIP~\cite{zhai2023siglip}, SAIL~\cite{zhang2025sail}, FLAIR~\cite{xiao2025flair}). By combining different encoders and VLMs, we provide the first large-scale comparative analysis of how backbone choice jointly affect CBM accuracy and interperability.

In summary, we make the following \textbf{contributions}:
\begin{itemize}
\item We introduce a principled, entropy-based metric to evaluate the relevance of concept sets in both task-agnostic and task-specific contexts, enabling pre-training assessment of concept quality.
\item We identify and address the linearity issue in CBMs, ensuring that predictions genuinely depend on concepts rather than degenerating into a linear probe.
\item We propose to use a knowledge distillation loss that reduces the performance gap between CBMs and the linear probe of the same encoder.
\item We conduct the first systematic study of how different vision encoders, and vision–language models 
interact to shape accuracy in CBMs.
\end{itemize}

\section{Related Work}

Concept Bottleneck Models (CBMs)~\citep{koh2020cbm, StammerSK21} aim to improve the interpretability of deep neural networks by enforcing a structured, human-understandable intermediate representation \citep{poeta2023concept}. Specifically, CBMs decompose the prediction process into two stages: a concept encoder $W_c$, which projects the input image $x \in \mathbb{R}^{H\times W \times 3}$ to a predefined set of interpretable concepts $\mathbb{C}$, and a classifier $W_f$, which predicts the final label $\hat{y}$ based solely on the concept activations $c \in \mathbb{R}^{K}$, where $K$ is the number of concepts. This architecture enables not only transparency in the model’s decision-making but also the possibility of human intervention, allowing users to inspect or edit concept activations to improve predictions. 

Early CBMs~\citep{koh2020cbm} relied on annotated attribute datasets for training, which are costly to obtain. Later, Post-hoc CBM (PCBM)~\citep{post_hoc_cbm} reduced the needed level of annotation for the dataset by facilitating the method called concept activation vectors (CAV)~\citep{kim2018cav}. More recent approaches~\citep{lfcbm, labo, sun2024leakage, srivastava2024vlgcbm} overcome this constraint by automatically extracting concepts using large language models (LLMs) with only class name supervision, thereby significantly expanding the usability and applicability of CBMs.

We classify existing CBMs into two main categories based on how their concept sets are derived: bottom-up and top-down approaches. Bottom-up CBMs derive their concepts directly from visual content, leveraging external models to extract or discover salient features (\eg, \citep{prasse2024dcbm, StammerWSK24}). DiscoverThenName~\cite{dncbm}, for instance, uses a sparse autoencoder to reconstruct CLIP embeddings and interprets the resulting basis vectors as concepts, which are later assigned to concept names by their cosine similarity. VLG-CBM~\citep{srivastava2024vlgcbm} adopts a similar visual grounding strategy, employing Grounding DINO~\citep{liu2023grdino} to construct a concept-labeled dataset directly from image regions. While effective for identifying visually grounded concepts, this approach is inherently limited in scope, as it cannot capture non-visual concepts, such as contextual cues or abstract, relational concepts, that are often critical for high-level understanding.

In contrast, top-down CBMs define their concept sets prior to training, typically relying on domain expertise or LLMs prompted with class-specific queries to extract descriptive attributes. Representative methods in this category include Post-hoc CBM~\citep{post_hoc_cbm}, LFCBM~\citep{lfcbm}, LaBo~\cite{labo}, and LM4CV~\citep{yan2023lm4cv}, all of which employ LLMs to generate concepts by querying with class names and tailored prompts. This approach offers greater flexibility, allowing multiple concept sets per dataset—for instance, a hematologist could define morphological features for cell classification. Ideally, CBMs should support such expert-driven, task-specific concept sets. In this work, we focus on developing a framework that advances the top-down CBM approach.
\section{Issues in Concept Bottleneck Models}
Let us now highlight several critical issues affecting recent (top-down) CBMs.

\subsection{Linearity Problem}
\label{sec:linearity_problem}

We identify a structural issue that affects many recent VLM-based CBMs~\citep{lfcbm,post_hoc_cbm,srivastava2024vlgcbm}. Specifically, when both the concept encoder and the classifier are implemented as linear transformations without any intermediate non-linearity, the model effectively collapses into a standard linear classifier that bypasses the concept bottleneck entirely.

For example, in LFCBM~\cite{lfcbm}, concept activations are computed as $W_cf(x)$, where $W_c$ is a learnable linear projection and $f(x)$ is the image embedding. A second linear layer $W_f$ is then trained on top of the concept activations, resulting in a final prediction of the form $W_fW_cf(x)$. Although the two stages are trained sequentially, the overall computation reduces to a single linear transformation $Wf(x)$, rendering the intermediate concept representation mathematically redundant and making the model's behavior independent of the concept set.

\begin{figure}[t!]
\centering
\begin{tikzpicture}
  \matrix[matrix of nodes, nodes={anchor=center}, column sep=1mm, row sep=1mm] {
    \parbox[t]{0.02\linewidth}{\rotatebox{90}{\small \textbf{Images}}} 
    &\includegraphics[height=0.30\linewidth]{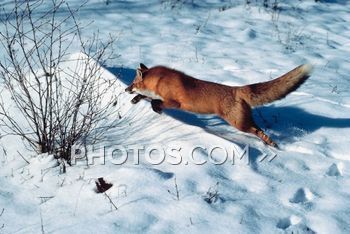}
    & \includegraphics[height=0.30\linewidth]{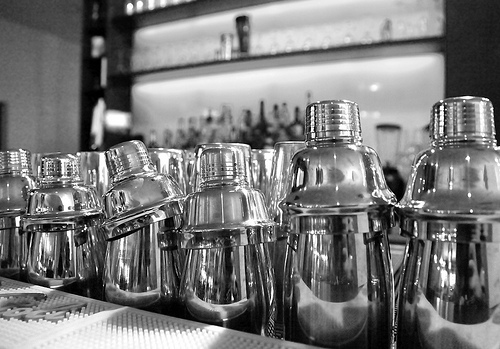} \\
    
    \parbox[t]{0.02\linewidth}{\rotatebox{90}{\small \textbf{Top-5 Concepts}}} 
    & \parbox[t][0.2\linewidth]{0.43\linewidth}{\small \textbf{1.} psybxtfoxqxlw \\ \textbf{2.} ukwnyhilccctox \\ \textbf{3.} hboxpfschan \\ \textbf{4.} yeoxs \\ \textbf{5.} mdacdbud} 
    & \parbox[t][0.2\linewidth]{0.43\linewidth}{\small  \textbf{1.} jurisprudentes \\ \textbf{2.} fifth \\ \textbf{3.} record \\ \textbf{4.} links \\ \textbf{5.} handle} \\
  };
\end{tikzpicture}
\caption{Linearity problem undermines interpretability. Linear CBMs attain high accuracy (85+\% on ImageNet100 dataset) with random strings (\textbf{left}) and Roman Law terminology (\textbf{right}).}
\label{fig:issues_lin}
\end{figure}

This degeneracy, which we refer to as the \textit{linearity problem}, directly undermines the foundational objective of CBMs: providing interpretable predictions grounded in human-understandable concepts (\cf \autoref{fig:issues_lin}). Notably, while early CBM work employed non-linear concept encoders to avoid this issue~\citep{koh2020cbm,StammerSK21}, recent VLM-based approaches have predominantly reverted to linear architectures, likely due to assumptions about the semantic richness of pretrained VLM embeddings or computational simplicity.

Our empirical findings (\autoref{fig:linearity}) confirm that under purely linear architectures, altering the concept set (\eg, replacing relevant concepts with random strings) has a negligible impact on classification accuracy, demonstrating that the model ignores the concept layer entirely. This reveals a critical gap in prior CBM evaluations, which have not systematically assessed model sensitivity to concept set choice. We argue that evaluating robustness to concept set variation—particularly through comparisons between relevant and irrelevant concept sets—should become standard practice in CBM research to ensure models genuinely leverage their concept bottlenecks.

\subsection{Accuracy Gap}
By design, CBMs constrain the prediction process to rely solely on the concept bottleneck layer. This architectural restriction, while enhancing interpretability, often results in a drop in predictive accuracy \cite{post_hoc_cbm, lfcbm, labo} relative to opaque end-to-end models.

To narrow this performance gap, previous studies have augmented the concept bottleneck with additional latent neurons, referred to as side-channel CBMs \cite{post_hoc_cbm, havasi2022leakage}. These auxiliary neurons allow the model to bypass the concept bottleneck and convey extra information to the final classifier. However, this undermines the interpretability of the model, as part of the prediction no longer depends exclusively on human-understandable concepts \cite{debot2025tradeoff}.

We address this issue by incorporating a knowledge distillation mechanism that leverages an opaque model as a teacher. This approach enables the concept-to-label mapping to benefit from the opaque model’s supervision while maintaining a fully interpretable bottleneck without introducing any non-interpretable neurons, thereby helping to close the accuracy gap.
\section{Method}
\label{sec:method}

\begin{figure*}[t]
\begin{center}
\includegraphics[width=0.9\textwidth]{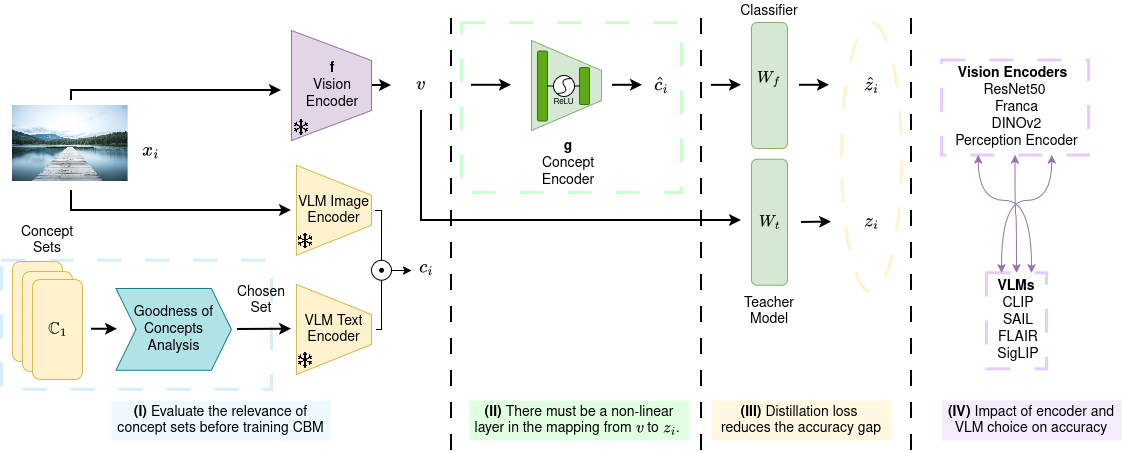}
\end{center}
\caption{Overview of the CBM-Suite, designed to address the key pitfalls of conventional Concept Bottleneck Models. First, the most relevant concept set for the dataset is selected using the Goodness of Concepts metric. Then, a non-linear concept encoder is trained to construct a trustworthy CBM. Next, the final classifier module is trained under teacher supervision. Finally, CBM-Suite provides a comprehensive evaluation suite across different vision encoders and vision-language models.}
\label{fig:pipeline}
\end{figure*}

Having identified critical limitations of existing CBMs, we now present \emph{CBM-Suite} and show how it addresses these. \autoref{fig:pipeline} illustrates the overall pipeline. We first define the Goodness of Concepts metric and describe its computation under two configurations: task-agnostic and task-specific. Subsequently, we introduce the Non-linear CBM architecture and formulate the knowledge distillation loss employed during training to enhance model performance.

\subsection{Evaluating Concept Set Relevance}
\label{sec:relevance-of-concepts}

We compute concept activations by measuring the cosine similarity between Vision Language Model's (VLM) image embeddings and the corresponding text embeddings of each concept. This procedure is applied independently for each concept set and dataset pair, allowing us to assess how well different sets align with the visual content across datasets. Formally, given an input image $x_i \in \mathbb{R}^{H\times W \times 3}$, its concept activations $ c_i \in \mathbb{R}^{K}$, where $K$ is the cardinality of the concept set, are calculated as:

\begin{equation}
    \displaystyle c_i = \dfrac{E_{text}(\mathbb{C}) \cdot E_{img}(x_i) -\mu}{\sigma},
    \label{eq:concept_act}
\end{equation}

\noindent where $E_{img}$ and $E_{text}$ denote any VLM's image and text encoders, respectively, and $\mathbb{C}$ is the concept set of size $K$. The image encoding $E_{img}(x_i)$ yields a $d\times 1$ latent feature vector, where $d$ is the feature dimension of the VLM. The text encoder generates the concatenated concept feature matrix $E_{text}(\mathbb{C}) \in \mathbb{R}^{K\times d}$, where each of the K rows is the d-dimensional embedding for a distinct concept in $\mathbb{C}$. \autoref{eq:concept_act} produces a vector of concept similarity scores per image. 
Let $\mu$ and $\sigma$ denote the mean and standard deviation of the concept activations computed over the training data. We normalize the concept activations by subtracting $\mu$ and dividing by $\sigma$.

\begin{figure}[t!]
\centering
\begin{subfigure}[b]{.45\linewidth}
\centering
\begin{tikzpicture}[scale=0.5, every node/.append style={transform shape}]
    \begin{axis}[
        title = ImageNet100,
        title style = {font=\Large},
        xtick={-0.4, -0.45,...,0.55},
        area style,
        ]
        \addplot+[draw=NavyBlue!80!black,
        fill=NavyBlue!60, fill opacity=0.7, ybar interval,mark=no] plot coordinates { 
            (-0.4, 0)
            (-0.35000000000000003, 0)
            (-0.30000000000000004, 0)
            (-0.25000000000000006, 0)
            (-0.20000000000000007, 0)
            (-0.15000000000000008, 0)
            (-0.10000000000000009, 0)
            (-0.0500000000000001, 0)
            (-1.1102230246251565e-16, 153)
            (0.04999999999999988, 1759720)
            (0.09999999999999987, 6538073)
            (0.1499999999999998, 3025437)
            (0.19999999999999984, 865529)
            (0.2499999999999999, 272127)
            (0.2999999999999998, 107218)
            (0.34999999999999976, 67862)
            (0.3999999999999998, 28560)
            (0.44999999999999984, 3626)
            (0.4999999999999998, 563)
            (0.5499999999999997, 32)

        };
        \addplot+[draw=Peach!80!black,
        fill=Peach!60, fill opacity=0.7, ybar interval,mark=no] plot coordinates { 
            (-0.4, 0)
            (-0.35000000000000003, 0)
            (-0.30000000000000004, 0)
            (-0.25000000000000006, 0)
            (-0.20000000000000007, 0)
            (-0.15000000000000008, 0)
            (-0.10000000000000009, 30)
            (-0.0500000000000001, 770399)
            (-1.1102230246251565e-16, 2077294)
            (0.04999999999999988, 3785925)
            (0.09999999999999987, 4079251)
            (0.1499999999999998, 1722910)
            (0.19999999999999984, 223756)
            (0.2499999999999999, 9237)
            (0.2999999999999998, 98)
            (0.34999999999999976, 0)
            (0.3999999999999998, 0)
            (0.44999999999999984, 0)
            (0.4999999999999998, 0)
            (0.5499999999999997, 0)
        };
    \end{axis}
\end{tikzpicture} 
\end{subfigure}
\begin{subfigure}[b]{.45\linewidth}
\centering
\begin{tikzpicture}[scale=0.5, every node/.append style={transform shape}]
    \begin{axis}[
        transform canvas={yshift=1cm, xshift=-2cm},
        title=Places365,
        title style = {font=\Large},
        xtick={-0.4, -0.45,...,0.55},
        area style,
        legend style={
        draw=none,
        at={(-0.2,-0.15)},
        anchor=north,
        legend columns=2,
        font=\Large,
        /tikz/every even column/.append style={column sep=0.3cm}
        },
        ]
        \addplot+[draw=NavyBlue!80!black,
        fill=NavyBlue!60, fill opacity=0.7, ybar interval,mark=no] plot coordinates {
            (-0.4, 0)
            (-0.35000000000000003, 0)
            (-0.30000000000000004, 0)
            (-0.25000000000000006, 0)
            (-0.20000000000000007, 0)
            (-0.15000000000000008, 0)
            (-0.10000000000000009, 0)
            (-0.0500000000000001, 0)
            (-1.1102230246251565e-16, 4876740)
            (0.04999999999999988, 81794113)
            (0.09999999999999987, 65700277)
            (0.1499999999999998, 16708255)
            (0.19999999999999984, 5448214)
            (0.2499999999999999, 3068652)
            (0.2999999999999998, 1512073)
            (0.34999999999999976, 672268)
            (0.3999999999999998, 296977)
            (0.44999999999999984, 167001)
            (0.4999999999999998, 85536)
            (0.5499999999999997, 15837)
            (0.5999999999999998, 57)
            };
        \addplot+[draw=Peach!80!black,
        fill=Peach!60, fill opacity=0.7, ybar interval,mark=no] plot coordinates {
            (-0.4, 0)
            (-0.35000000000000003, 0)
            (-0.30000000000000004, 0)
            (-0.25000000000000006, 0)
            (-0.20000000000000007, 0)
            (-0.15000000000000008, 0)
            (-0.10000000000000009, 144)
            (-0.0500000000000001, 4080057)
            (-1.1102230246251565e-16, 17958980)
            (0.04999999999999988, 39731633)
            (0.09999999999999987, 61308771)
            (0.1499999999999998, 47393951)
            (0.19999999999999984, 9326462)
            (0.2499999999999999, 535981)
            (0.2999999999999998, 9977)
            (0.34999999999999976, 44)
            (0.3999999999999998, 0)
            (0.44999999999999984, 0)
            (0.4999999999999998, 0)
            (0.5499999999999997, 0)
            };
    \legend{Relevant Concept Set, Irrelevant Concept Set}
    \end{axis}
\end{tikzpicture}
\end{subfigure}
\vspace{0.3cm}
\caption{Histogram of concept activations computed with SAIL for relevant and irrelevant concept sets on ImageNet100 and Places365, illustrating the difference in activation distributions between meaningful and random concepts.}
\label{fig:concept_act_dist_sail}
\end{figure}
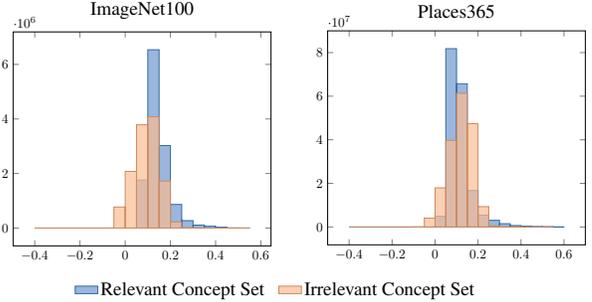

While the overall activation statistics remain similar across different configurations, it is essential to examine their distributional properties to understand finer-grained differences in how the concept set aligns with the dataset. We first visualize the activation distributions in \autoref{fig:concept_act_dist_sail}, highlighting the differences between relevant and irrelevant concept sets. For both ImageNet100 and Places365, the activations corresponding to irrelevant concepts follow an approximately normal distribution. In contrast, the distribution of relevant concepts is noticeably narrower, reflecting more concentrated activation pattern.

Under the assumption that effective explanations should be sparse \citep{wong2021sparse}, to analyze the relevance of the concepts, we first convert concept activations into probability distributions with the softmax function and then apply the standard entropy formulation over the resulting distribution. In this way, lower entropy values correspond to more concentrated and interpretable explanations. Importantly, we assess this \textbf{Goodness of Concepts} metric both at the dataset level (task-agnostic), and at the class level (task-specific). 
The distinction between these two settings stems from the observation that certain concepts may be present in an image yet remain irrelevant to the downstream classification task. While the task-agnostic goodness of concepts reflects the overall presence of concepts, task-specific goodness of concepts accounts for their relevance to the target labels by averaging activations over samples of the same class.



\textbf{Task-agnostic Goodness of Concepts.} In the task-agnostic setting, we evaluate concept quality by computing activations for all images in the dataset, independent of class labels (hence, task agnostic). We apply the softmax function to obtain a valid probability distribution from each activation vector $c_i$:
\begin{equation}
   c_{i,k}^\prime = \mathrm{softmax}( c_{i,k}) = \dfrac{\exp( c_{i,k})}{\sum_{j\in [1, K]}  \exp( c_{i,j})}.
   \label{eq:softmax}
\end{equation}
We then compute the entropy of the probability distribution:
\begin{equation}
    H(c_i^\prime) = -\sum_{j\in [1,K]} c^\prime_{ij}\log c^\prime_{ij}.
    \label{eq:entropy}
\end{equation}

The average entropy across images is reported as the measure of the task-agnostic goodness of concepts. Lower entropy values indicate more selective activations, suggesting that only a few concepts are highly responsive to a given image, an attribute desirable for interoperability.

\textbf{Task-specific Goodness of Concepts.}
Given a dataset $D\in \mathbb{R}^{N\times H\times W \times 3}$ of size $N$, we extract class-specific subsets $D_s\subset D$, where each subset contains $n$ samples belonging to class $s$. For each class, we compute the average concept activations by aggregating the cosine similarities between the VLM's image embeddings and the text embeddings of the concept set $\mathbb{C}$. Formally, the average concept activation vector $c_{s}\in \mathbb{R}^K$ for class $s$ is computed as:

\begin{equation}
    c_s =  \frac{1}{n} \sum_{x_i\in D_s} \dfrac{E_{text}(\mathbb{C}) \cdot E_{img}(x_i) -\mu}{\sigma}.
\end{equation}

We then normalize $c_s$ using the softmax function, as shown in \autoref{eq:softmax}, to obtain $c_s^\prime = \text{softmax}(c_s)$. Subsequently, we compute the entropy per class, $H(c_s^\prime)$, following \autoref{eq:entropy}, and report the average entropy across all classes. The task-specific goodness of concepts metric quantifies how selectively a concept set captures information relevant to each class and provides a valuable measure for distinguishing between concept sets that genuinely capture task-relevant semantics and those susceptible to concept leakage.

\subsection{Non-linear Concept Encoding}
\label{sec:concept-encoder}

To address the linearity problem identified in recent VLM-based CBMs, we employ a non-linear concept encoder that maps image embeddings to the concept space. While early concept bottleneck work utilized non-linear encoders \citep{StammerSK21}, recent VLM-based approaches have largely adopted linear mappings from image embeddings to concepts. However, as discussed in \autoref{sec:linearity_problem}, this architectural choice causes the model to degenerate into a linear probe that bypasses the concept bottleneck entirely -- the overall computation reduces to a single linear transformation, rendering the intermediate concept representation mathematically redundant.

We employ a two-layer multilayer perceptron with ReLU activation to map image features $f(x_i)$ extracted from the vision backbone into the concept space. The target concept activations $c_i$ are obtained from the VLM as described in \autoref{sec:relevance-of-concepts}. Letting $\hat{c}_i$ represent the predicted concept activations, we train the concept encoder using the  mean squared error loss:
\begin{equation}
    L_{CE} = \dfrac{1}{N}\sum_{i=1}^N \|\hat c_{i} - c_{i}\|^2.
\end{equation}

This simple non-linearity introduced by the ReLU activation ensures that the model cannot bypass the concept bottleneck through matrix composition, forcing predictions to genuinely pass through the concept space.

\subsection{Teacher-Guided Training}
\label{sec:teacher-distillation}

To address the accuracy gap between CBMs and standard classifiers, we introduce a knowledge distillation framework where a linear probe trained directly on image embeddings serves as the teacher model. This approach allows the CBM to recover the predictive performance of an unconstrained classifier while maintaining interpretability through the concept bottleneck.

\textbf{Teacher Model.}  
The teacher model is implemented as a linear probe trained on the image embeddings $f(x_i)$ extracted from the vision backbone. With weight matrix $W_t$, its predictions are $W_t f(x_i)$, supervised using the cross-entropy loss with respect to the ground-truth labels $y_i$:
\begin{equation}
    L_{T} = -\sum_{i=1}^N y_i \log (\mathrm{softmax}(W_t f(x_i))).
\end{equation}

\textbf{Final Classifier with Knowledge Distillation.}  
On top of the learned concept activations, we train a final classification layer with weight matrix $W_f$. The classifier loss combines three components:
\begin{itemize}
    \item \textbf{Cross-entropy loss} computed against the ground-truth labels $y_i$
    \item \textbf{Elastic net regularization} implemented via $\ell_1$ and $\ell_2$ penalties on $W_f$ to encourage sparse, interpretable concept usage
    \item \textbf{Knowledge distillation} from the teacher model, aligning the student logits ($W_f \hat c_i$) with the teacher logits ($ z_i = W_t f(x_i)$)
\end{itemize}

The total loss is expressed as a weighted sum of these terms:
\begin{equation}
\begin{split}
    L_C =& \alpha \Bigg[\overbrace{-\sum_{i=1}^N y_i \log( \mathrm{softmax}(W_f \hat c_{i}))}^{\text{Cross-entropy}} \\
    &\qquad + \overbrace{\vphantom{\sum_{i=1}^N}\lambda\|W_f\|_1 + (1-\lambda)\|W_f\|^2}^{\text{Elastic net regularization}} \Bigg] \\
         & + \beta \Bigg[ \underbrace{T^2  \sum_{i=1}^N z_{i}\big(\log (\dfrac{W_f \hat c_{i}}{T}) - \log \dfrac{ z_{i}}{T}\big)}_{\text{Knowledge distillation}} \Bigg],
\end{split}
\label{eq:classifier_loss}
\end{equation}

\noindent where $\alpha$ and $\beta$ are weighting coefficients balancing the supervised and distillation objectives, $T$ denotes the temperature used for distillation, and $\lambda$ controls the balance between $\ell_1$ and $\ell_2$ regularization. 

\textbf{Summary.} In summary, CBM-Suite introduces a set of complementary techniques for addressing key limitations of recent CBM approaches: (i) an entropy-based Goodness of Concepts metric for evaluating concept set relevance prior to training; (ii) a non-linear concept encoder that ensures predictions genuinely pass through the concept bottleneck rather than degenerating to a linear probe; and (iii) knowledge distillation from a teacher model to close the accuracy gap between CBMs and standard classifiers.

\section{Experiments}
\label{experiments}

\textbf{Experimental Setup.}
We evaluate the CBM-Suite on four datasets: ImageNet100~\citep{IN}, Places365~\citep{places}, CUB200~\citep{cub}, CIFAR100~\citep{cifar}. For relevant concept sets, we adopt those generated by LFCBM~\citep{lfcbm}, which uses GPT3 to generate concepts based on class names. Since most existing CBM literature employs ResNet as the vision backbone, we follow the same convention for consistency while evaluating the CBM-Suite for the first 3 issues. Specifically, ResNet-50 is used for ImageNet100 and Places365, while ResNet-18 pretrained and fine-tuned on CUB200 is used for that dataset. 

The Non-linear CBM is optimized using the Adam optimizer, along with a learning rate scheduler. The dataset-specific learning rates and scheduler configurations are provided in the supplementary material \autoref{sup:configurations}.

\textbf{Goodness of Concepts.} We assess the suitability of various concept sets across different datasets, as described in \autoref{sec:relevance-of-concepts}. Specifically, we compute concept activations for each image in the ImageNet100, CUB200, and Places365 datasets using three distinct concept sets: (1) a relevant concept set generated by a large language model (GPT3~\cite{GPT3}), as motivated by prior studies \cite{lfcbm, labo} for ImageNet100 and Places365, and human-annotated attributes for CUB200, (2) an irrelevant concept set derived from Roman Law terminology, and (3) a baseline set of random strings.

\begin{table}[t!]
\centering
\caption{\textbf{Goodness of concepts}: suitability of different concept sets for different datasets. Relevant concepts are GPT-generated in ImageNet100 (IN100) and Places365 and GT attributes in CUB200. The irrelevant concept set consists of Roman Law terminology. Random strings denote non-word concepts. \textbf{Bold}/\underline{underline} indicate lowest/highest entropy (best/worst).}
\label{tab:goodness_concepts}
\vspace{0.5em}
\resizebox{0.83\linewidth}{!}{%
\begin{tabular}{@{}l*{3}{c}@{}}
\toprule
\multicolumn{4}{c}{\textbf{Task Agnostic} $\downarrow$} \\
\cmidrule(lr){1-4}
\textbf{Concept Set} & IN100 & CUB200 & Places365 \\
\midrule
Relevant  & \textbf{3.660} & \textbf{4.305} & \textbf{3.077} \\
Irrelevant & 4.504 & 4.491 & 4.505 \\
Random words & \underline{4.565} & \underline{4.553} & \underline{4.566} \\
\midrule
\multicolumn{4}{c}{\textbf{Task Specific} $\downarrow$} \\
\cmidrule(lr){1-4}
\textbf{Concept Set} & IN100 & CUB200 & Places365 \\
\midrule
Relevant  & \textbf{3.738} & \textbf{4.383} & \textbf{2.889}\\
Irrelevant & 4.545 & 4.526 & 4.547 \\
Random words & \underline{4.579} & \underline{4.565} & \underline{4.578}\\
\bottomrule
\end{tabular}
}
\end{table}

For the irrelevant concept set, we retrieved Roman Law content from the relevant Wikipedia page. The resulting text was processed with a natural language processing (NLP) pipeline \cite{spacy}, from which we extracted nouns, adjectives, and verbs. We then removed semantically redundant terms to construct a compact, non-overlapping concept set. Next, we created a random concept set of size 2,500 by sampling lowercase alphabetic strings of random lengths uniformly drawn from [5, 15).

In \autoref{tab:goodness_concepts}, we report the entropy values in two settings: task-agnostic and task-specific. First, we compute the concept activations for each image using SAIL~\cite{zhang2025sail} as the VLM (due to its wider similarity distribution compared to CLIP -- see supplementary material \autoref{sup:vlm_dist}). 
Since entropy depends on the size of the concept set, we introduce a cutoff on the concept activations to ensure a fair comparison across concept sets of different cardinalities. We note that the goodness of concepts metric remains robust across varying cutoff values (see supplementary material \autoref{sup:cutoff_values}).
In the task-agnostic setting, we identify the top-100 activated concepts for each image, compute the entropy over these activations, and report the average across the dataset. In the task-specific setting, we first average concept activations per class to identify the top-100 class-specific concepts before computing the corresponding entropy. This analysis provides insight into the discriminative and descriptive capacity of concept sets.

As a measure of randomness, entropy quantifies the importance of careful concept set design. Lower entropy values indicate more concentrated and potentially semantically meaningful activations, reflecting a stronger alignment between the concept set and the visual domain. Among all sets, relevant concept sets consistently achieve the lowest entropy across nearly all configurations (see supplementary material \autoref{sup:GoC}). 
As expected, random strings and Roman Law concepts, which are semantically unrelated to the image domains, yield the highest entropy, indicating weak alignment with the visual features. 
Preliminary experiments suggest that our goodness of concepts metric is effective to refine a concept set by finding irrelevant concepts to remove. (see supplementary material \autoref{sup:refining}).

\pgfplotstableread{
0 90.37 90.58 0.22 0.22 0.23 0.23
1 88.45 62.38 0.45 0.45 8.36 8.36
}\indataset

\pgfplotstableread{
0 40.32 40.11 0.86 0.86 0.88 0.88
1 35.44 28.63 1.42 1.42 4.47 4.47
}\cubdataset

\begin{figure}[t!]
\centering
\begin{subfigure}[b]{\linewidth}
\centering
\begin{tikzpicture}
\begin{axis}[ybar,
        title = ImageNet100,
        height=4cm,
        width=\linewidth,
        bar width=15,
        xmin=-0.5,xmax=1.5,
        ymin=45,
        ymax=95,
        ylabel={Accuracy (\%)},
        ylabel near ticks,
        ylabel style={font=\small\bfseries},
        xtick=data,
        xtick={ 0,1},
        xticklabels={
            Linear CBM,
            Non-linear CBM
        },
        major x tick style = {opacity=0},
        ymajorgrids=true,
        grid style={dashed, gray!30},
        legend style={draw=none}
        ]
    \addplot [Mahogany,line legend,
        sharp plot,update limits=false,
        ] coordinates { (-1 ,91.3) (2,91.3) };
    \addplot[draw=BrickRed!90!black,
        fill=BrickRed!70,
        error bars/.cd,
        y dir=both,
        y explicit,
        error bar style={line width=1pt, BrickRed!100!black}] 
    table[x index=0,y index=1,y error plus index=3,y error minus index=4] \indataset; 
    \addplot[draw=Dandelion!60!black,
        fill=Dandelion!40,
        error bars/.cd,
        y dir=both,
        y explicit,
        error bar style={line width=1pt, Dandelion!70!black}] 
    table[x index=0,y index=2,y error plus index=5,y error minus index=6] \indataset; 
\end{axis}
\end{tikzpicture}
\end{subfigure}
\vspace{0.1cm}
\begin{subfigure}[b]{\linewidth}
\centering
\begin{tikzpicture}
\begin{axis}[ybar,
        title = Places365,
        height=4cm,
        width=\linewidth,
        bar width=15,
        xmin=-0.5,xmax=1.5,
        ymin=10,
        ymax=60,
        ylabel={Accuracy (\%)},
        ylabel near ticks,
        ylabel style={ font=\small\bfseries},
        xtick=data,
        xtick={ 0,1},
        xticklabels={
            Linear CBM,
            Non-linear CBM
        },
        major x tick style = {opacity=0},
        ymajorgrids=true,
        grid style={dashed, gray!30},
        legend style={draw=none, 
        at={(0.5,-0.4)},
        anchor=north,
        legend columns=3,
        font=\small,
        /tikz/every even column/.append style={column sep=0.3cm}
        }
        ]
    \addplot [Mahogany,line legend,
            sharp plot,update limits=false,
        ] coordinates { (-1 ,47.33) (2,47.33) };
    \addplot[draw=BrickRed!90!black,
        fill=BrickRed!70,
        error bars/.cd,
        y dir=both,
        y explicit,
        error bar style={line width=1pt, BrickRed!100!black}] 
    table[x index=0,y index=1,y error plus index=3,y error minus index=4] \cubdataset; 
    \addplot[draw=Dandelion!60!black,
        fill=Dandelion!40, 
        error bars/.cd,
        y dir=both,
        y explicit,
        error bar style={line width=1pt, Dandelion!70!black}] 
    table[x index=0,y index=2,y error plus index=5,y error minus index=6] \cubdataset;
\legend{Oracle, Relevant, Irrelevant}
\end{axis}
\end{tikzpicture}
\end{subfigure}    
\caption{\textbf{Effect of non-linear concept encoders.} We report classification accuracy on ImageNet100 and Places365 using both relevant and irrelevant concept sets. Linear CBM and Non-linear CBM differ only in the concept encoder layer, \ie, the latter includes a non-linear function. Error bars represent the standard deviation over ten independent runs with random initializations.}
\label{fig:linearity}
\end{figure}
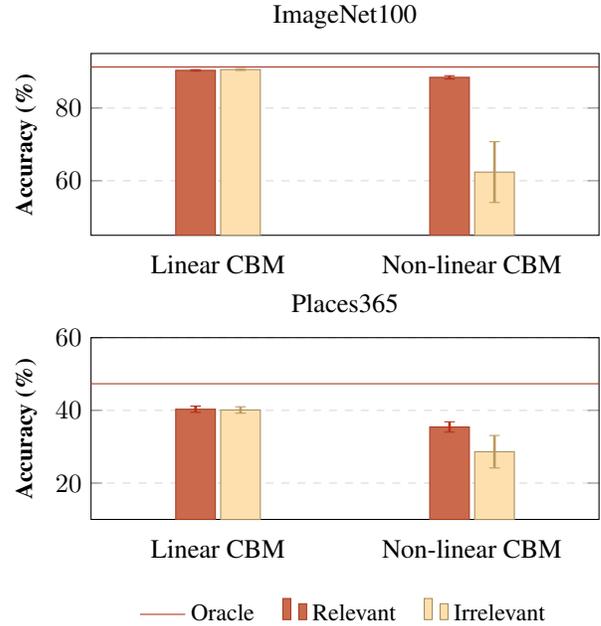

\textbf{Linearity.} To investigate the linearity issue in CBMs, we evaluate the performance of the Non-linear CBM, introduced in \autoref{sec:concept-encoder}, trained on the ImageNet100 and Places365 datasets using two concept sets: GPT-3–generated (relevant) and random words (irrelevant). We compare it to a Linear CBM obtained by removing the non-linearity function from the concept encoder. Both employ SAIL~\cite{zhang2025sail} encoders to compute the concept activations. Additionally, we report the classification accuracy of a linear probe trained directly on ResNet image embeddings, indicated as Oracle in \autoref{fig:linearity}, which serves as a baseline for purely feature-driven classification.

In \autoref{fig:linearity}, a distinct behavioral difference is observed between the Non-linear and Linear CBMs. The Linear CBMs remain unaffected by the choice of concept set, achieving high accuracy even with non-word concepts. This observation highlights that the absence of a non-linear function in the concept encoder can compromise the trustworthiness of the CBM, as it may rely on simple linear transformations rather than meaningful concept representations. 

On the other hand, the Non-linear CBMs perform significantly worse when trained with the irrelevant concept set. For example, on the ImageNet100 dataset, accuracy drops by roughly 25 points when trained with the irrelevant concept set. Moreover, the larger error bars for Non-linear CBMs with irrelevant concept set across both datasets further suggest that their performances are unstable and exhibit high variance under these condition.

\pgfplotstableread{
0 91.3 85.47 82.27 0.12 0.12 0.38 0.38 0.74 0.74 
}\distilldatasetIN

\pgfplotstableread{
0 47.33 35.44 33.64 0.10 0.10 1.42 1.42 0.50 0.50
}\distilldatasetPlaces

\begin{figure}[t!]
\centering
\begin{subfigure}[b]{.45\linewidth}
\centering
\begin{tikzpicture}
    \begin{axis}[
    title = ImageNet100,
    ybar=2pt,
    bar width=12pt,
    width=\linewidth,
    height=5cm,
    xmin=-0.5, xmax=0.5,
    ymin=80,
    ymax=94,
    ylabel={Accuracy (\%)},
    ylabel near ticks, 
    ylabel style={font=\small\bfseries},
    xtick=data,
    xtick={0},
    xticklabel style={opacity=0},
    yticklabel style={font=\small},
    major x tick style={opacity=0},
    ymajorgrids=true,
    grid style={dashed, gray!30},
    legend style={
        draw=none,
        at={(0.5,-0.25)},
        anchor=north,
        legend columns=3,
        font=\small,
        /tikz/every even column/.append style={column sep=0.3cm}
    },
    legend cell align={left},
    enlarge x limits=0.25,
    ]
    
    \addplot[
        draw=Gray!70,
        fill=Gray!50,
        pattern=north east lines,
        pattern color=black!40,
        error bars/.cd,
        y dir=both,
        y explicit,
        error bar style={line width=1pt, Gray!80!black}
    ] table[x index=0,y index=1,y error plus index=4,y error minus index=5] \distilldatasetIN;
    \addplot[
        draw=SeaGreen!60!black,
        fill=SeaGreen!40,
        pattern=north east lines,
        pattern color=SeaGreen!80,
        error bars/.cd,
        y dir=both,
        y explicit,
        error bar style={line width=1pt, SeaGreen!80!black}
    ] table[x index=0,y index=3,y error plus index=8,y error minus index=9] \distilldatasetIN;
    \addplot[
        draw=SeaGreen!60!black,
        fill=SeaGreen!40,
        error bars/.cd,
        y dir=both,
        y explicit,
        error bar style={line width=1pt, SeaGreen!80!black}
    ] table[x index=0,y index=2,y error plus index=6,y error minus index=7] \distilldatasetIN;
    \end{axis}
\end{tikzpicture} 
\end{subfigure}\hfill
\begin{subfigure}[b]{.45\linewidth}
\centering
\begin{tikzpicture}
    \begin{axis}[
    transform canvas={yshift=0.6cm, xshift=-1cm},
    title = Places365,
    ybar=2pt,
    bar width=12pt,
    width=\linewidth,
    height=5cm,
    xmin=-0.5, xmax=0.5,
    ymin=25,
    ymax=55,
    ylabel={Accuracy (\%)},
    ylabel near ticks, 
    ylabel style={font=\small\bfseries},
    xtick=data,
    xtick={0},
    xticklabel style={opacity=0},
    yticklabel style={font=\small},
    major x tick style={opacity=0},
    ymajorgrids=true,
    grid style={dashed, gray!30},
    legend style={
        draw=none,
        at={(-0.5,-0.15)},
        anchor=north,
        legend columns=3,
        font=\small,
        /tikz/every even column/.append style={column sep=0.3cm}
    },
    legend cell align={left},
    enlarge x limits=0.25,
    ]
    \addplot[
        draw=Gray!70,
        fill=Gray!50,
        pattern=north east lines,
        pattern color=black!40,
        error bars/.cd,
        y dir=both,
        y explicit,
        error bar style={line width=1pt, Gray!80!black}
    ] table[x index=0,y index=1,y error plus index=4,y error minus index=5] \distilldatasetPlaces;
    \addplot[
        draw=SeaGreen!60!black,
        fill=SeaGreen!40,
        pattern=north east lines,
        pattern color=SeaGreen!80,
        error bars/.cd,
        y dir=both,
        y explicit,
        error bar style={line width=1pt, SeaGreen!80!black}
    ] table[x index=0,y index=3,y error plus index=8,y error minus index=9] \distilldatasetPlaces;
    \addplot[
        draw=SeaGreen!60!black,
        fill=SeaGreen!40,
        error bars/.cd,
        y dir=both,
        y explicit,
        error bar style={line width=1pt, SeaGreen!80!black}
    ] table[x index=0,y index=2,y error plus index=6,y error minus index=7] \distilldatasetPlaces;
    
    \legend{Oracle, Vanilla CBM, \textbf{Distilled CBM}}
    \end{axis}
\end{tikzpicture}
\end{subfigure}
\vspace{0.3cm}
\caption{\textbf{Effect of knowledge distillation on CBM performance.} Classification accuracy on ImageNet100 and Places365 datasets. Distilled CBM significantly outperforms Vanilla CBM by incorporating distillation loss alongside the elastic loss, approaching Oracle performance. Error bars represent the standard deviation over ten independent runs  with random initializations.}
\label{fig:distillation}
\end{figure}
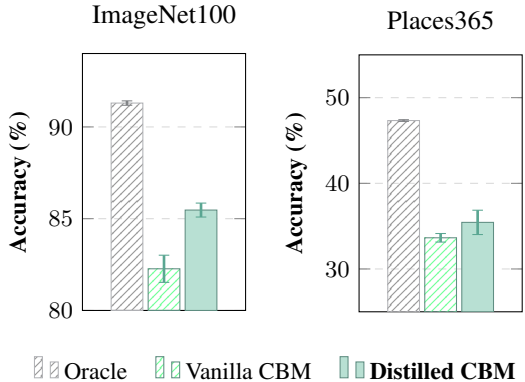

\textbf{Knowledge Distillation.}
We conduct an ablation study to examine the effect of the knowledge distillation term. In \autoref{fig:distillation}, we present the results for the Distilled CBM and the Vanilla CBM. The Distilled CBM is trained using the full objective described in \autoref{eq:classifier_loss}, while the Vanilla CBM is optimized only with the first term of this loss. For reference, we also report the performance of a linear probe trained directly on the backbone model’s embeddings, denoted as Oracle. During the training of the Distilled CBM, this Oracle model serves as the teacher.

Knowledge distillation consistently improves classification accuracy across datasets. On ImageNet100, the Distilled CBM achieves 85.47\%, outperforming the Vanilla CBM (82.27\%) and narrowing the gap toward the Oracle (91.30\%). Similarly, on Places365, the Distilled CBM reaches 35.44\%, compared to 33.64\% for the Vanilla CBM and 47.33\% for the Oracle. These results demonstrate that incorporating the distillation loss enables the CBM to better leverage the teacher model’s supervision, leading to improved classification performance.

\textbf{Effects of Different Model Components.}
The modular structure of CBMs enables the flexible integration and evaluation of various encoders. \autoref{tab:components} highlights how vision backbone strength and concept alignment jointly influence model performance. We are reporting the classification accuracies obtained using different combinations of vision backbones and VLMs on the ImageNet100 dataset. The results reveal a consistent performance improvement with stronger visual representations. 

Perception Encoder~\cite{bolya2025pe} achieves the highest accuracy across all vision backbones, reaching up to 96.56\% with the linear probe baseline, indicating its strong representational capacity. Across VLMs, CLIP~\cite{clip} and SigLIP~\cite{zhai2023siglip} consistently outperform others. 
These findings emphasize that both the choice of backbone and the VLM substantially affect CBM performance.

\begin{table}[t!]
\centering
\caption{\textbf{Backbone choice matters more than VLM selection.} Performance accuracies (\%) of different VLM and vision backbones on ImageNet100 dataset. Perc.Enc. consistently outperforms other backbones across all methods with performance generally varying more across backbones (rows) than VLMs (columns). SigLIP performs best among VLMs in 3 of 4 cases, with the Perc.Enc. + SigLIP combination yielding best overall performance.}
\label{tab:components}
\vspace{0.5em}
\resizebox{0.98\linewidth}{!}{%
\begin{tabular}{@{}l c | *{4}{c}@{}}
\toprule
\textbf{Backbone} & \textbf{LP} & \multicolumn{4}{c}{\textbf{VLM}} \\
\cmidrule(lr){3-6}
 & & \textbf{CLIP}~\cite{clip} & \textbf{SAIL}~\cite{zhang2025sail} & \textbf{FLAIR}~\cite{xiao2025flair} & \textbf{SigLIP}~\cite{zhai2023siglip} \\
\midrule
ResNet~\cite{he2016resnet}              & 91.30 & 89.16 & 85.56 & 87.00 & 90.12 \\
DINOv2~\cite{caron2021dinov2}           & 92.72 & 92.22 & 89.28 & 89.68 & 91.74 \\
Franca~\cite{venkataramanan2025franca}  & 94.10 & 93.66 & 88.40 & 90.08 & 93.68 \\
Perc. Enc.~\cite{bolya2025pe}           & 96.56 & 95.84 & 93.56 & 93.84 & \textbf{96.14} \\
\bottomrule
\end{tabular}
}
\end{table}

\textbf{Comparison with the SOTA CBMs.}
In \autoref{tab:comparison}, we compare our results with LaBo~\citep{labo} and LFCBM~\citep{lfcbm}, using the values reported in their respective papers. Our results obtained using the best-performing vision backbone–VLM pairs for each dataset. For CIFAR100, Places365, and ImageNet, we use a Perception Encoder as the backbone model, while for CUB200 we use ResNet18. CBM-Suite achieves the best performance on CUB200, CIFAR100, and Places365, and remains competitive on ImageNet.

\begin{figure}[t!]
\centering
\begin{tikzpicture}
  \matrix[matrix of nodes, nodes={anchor=center}, column sep=0.5mm, row sep=1mm] {
     & \parbox{0.2\linewidth}{\centering\footnotesize Class Name} 
     & \parbox{0.2\linewidth}{\centering\footnotesize Top-5 Concepts} 
     &
     & \parbox{0.2\linewidth}{\centering\footnotesize Class Name} 
     & \parbox{0.2\linewidth}{\centering\footnotesize Top-5 Concepts}\\
     
    \parbox[t][0.20\linewidth]{0.01\linewidth}{\rotatebox{90}{\centering\footnotesize ImageNet-100}} 
    &  \parbox[t][0.20\linewidth]{0.2\linewidth}{\centering\tiny Carbonara \\ \includegraphics[height=0.8\linewidth]{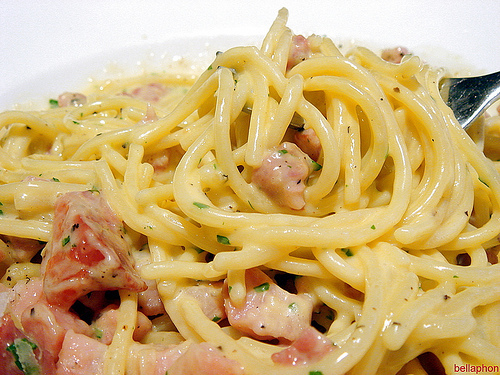}} 
    & \parbox[t][0.20\linewidth]{0.2\linewidth}{\tiny 1. pasta \\ 2. bacon or pancetta \\ 3. Italian dish \\ 4. noodles \\ 5. a creamy sauce} 
    & \parbox[t][0.20\linewidth]{0.01\linewidth}{\rotatebox{90}{\centering\footnotesize CUB200}} 
    &  \parbox[t][0.20\linewidth]{0.2\linewidth}{\centering\tiny Purple Finch \\ \includegraphics[height=0.8\linewidth]{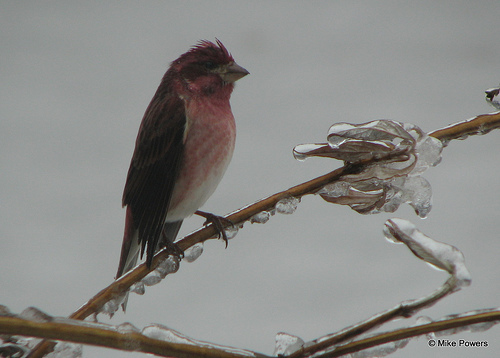}}
    & \parbox[t][0.20\linewidth]{0.2\linewidth}{\tiny 1. a red head \\ 2. a rosy breast \\ 3. a pinkish breast \\ 4. a purple-red head and breast \\ 5. a white eye-ring} \\
    
    \parbox[t][0.20\linewidth]{0.01\linewidth}{\rotatebox{90}{\centering\footnotesize CIFAR100}} 
    &  \parbox[t][0.20\linewidth]{0.2\linewidth}{\centering\tiny Telephone \\ \includegraphics[height=0.8\linewidth]{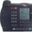}}
    & \parbox[t][0.20\linewidth]{0.2\linewidth}{\tiny 1. a phone \\ 2. communication device \\ 3. a keypad on one side \\ 4. a phone book \\ 5. lateral symmetry} 
    & \parbox[t][0.20\linewidth]{0.01\linewidth}{\rotatebox{90}{\centering\footnotesize Places365}} 
    & \parbox[t][0.20\linewidth]{0.2\linewidth}{\centering\tiny Palace \\ \includegraphics[height=0.8\linewidth]{qualitative/palace.png}} 
    & \parbox[t][0.20\linewidth]{0.2\linewidth}{\tiny 1. a royal residence \\ 2. grand, ornate architecture \\ 3. often has towers or turrets \\ 4. guards \\ 5. a garden on top of a building } \\
    };
\end{tikzpicture}
\caption{\textbf{Top concepts are well-matched with their respective classes.} Illustration of the Top-5 concepts, ranked by their contribution to the final prediction for randomly selected classes, along with representative images from ImageNet, CIFAR100, CUB200, and Places365, demonstrating that the selected backbone–VLM pairs capture meaningful semantic structure across domains.}
\label{fig:top5concepts}
\end{figure}

Finally, \autoref{fig:top5concepts} presents the top activated concepts for randomly selected classes across datasets, corresponding to the models reported in \autoref{tab:comparison}. If we look at the concepts closely, we observe that they are well-matched with the respective classes, further illustrating that the selected backbone–VLM pairs capture meaningful semantic structure. An extended version of this figure is provided in the supplementary material \autoref{sup:qualitative}.

\begin{table}[t!]
\centering
\caption{\textbf{Quantitative comparison with state-of-the-art CBMs.} Results for LaBo and LFCBM are taken from their respective papers. CBM-Suite uses the best-performing vision backbone–VLM pair for each dataset.}
\label{tab:comparison}
\vspace{0.5em}
\resizebox{0.98\linewidth}{!}{%
\begin{tabular}{@{}l | *{4}{c}@{}}
\toprule
\textbf{Models} & \textbf{CUB200}~\citep{cub} & \textbf{CIFAR100}~\citep{cifar} & \textbf{Places365}~\citep{places}  & \textbf{ImageNet}~\citep{IN}\\
\midrule
LaBo            & 81.00\% & 86.82\% & - & \textbf{83.97\%} \\
LFCBM           & 74.31\% & 65.13\% & 43.68\% & 71.95\% \\
CBM-Suite      & \textbf{86.73\%} & \textbf{92.50\%} & \textbf{54.64\%} & 81.56\% \\ 
\bottomrule
\end{tabular}
}
\end{table}

\section{Discussion}

Our findings highlight several important considerations for the development and evaluation of Concept Bottleneck Models. First, the goodness of concepts cannot be judged with a single global threshold. Concept activation distributions vary across different VLMs. Consequently, the quality and relevance of a concept set must be assessed per VLM rather than assumed to be universal. Second, although we demonstrate that introducing non-linearity in the concept encoder is crucial to overcome the linearity problem, it also comes with a measurable performance drop. However, we show that this trade-off can be mitigated through knowledge distillation, enabling CBMs to retain their predictive performance. Finally, while distilled CBMs produce top activated concepts that appear highly relevant, these models need to be further investigated against concept leakage as well. Future research must investigate precisely how distilled CBMs rely on concepts and determine whether they inadvertently encode label information outside the declared concept set.
Overall, these considerations underscore the need for nuanced, model-aware evaluation and continued refinement to ensure CBMs are both interpretable and robust.
\section{Conclusion}
\label{sec:conc}

In this work, we critically revisited Concept Bottleneck Models to reveal fundamental pitfalls undermining their interpretability and reliability. Through CBM-Suite, our comprehensive assessment framework, we demonstrate that concept set quality is central to meaningful explanations. Our entropy-based metric effectively distinguishes relevant concepts from irrelevant ones, addressing a key evaluation gap.
We found that linear CBM designs paradoxically achieve high accuracy even with irrelevant concepts, suggesting superficial rather than genuine interpretability. Incorporating non-linearity into the concept encoder strengthens the alignment between image features and concepts, while knowledge distillation substantially enhances performance, closing the gap to opaque baselines. Our evaluations across diverse vision backbones and VLMs reveal how representational strength directly influences accuracy. By systematically addressing these design choices, we provide practical guidance for developing CBMs that balance high performance with interpretability.
\section*{Acknowledgements}
We gratefully acknowledge the computational resources provided by METU-ROMER, Center for Robotics and Artificial Intelligence.
Dr. Akbas gratefully acknowledges the support of TUBITAK 2219. This work was partially funded by the ERC (853489 - DEXIM) and the Alfried Krupp von Bohlen und Halbach Foundation, for which we thank them for their generous support.

\newpage
{
    \small
    \bibliographystyle{ieeenat_fullname}
    \bibliography{main}

@String(PAMI  = {IEEE TPAMI})

@String(CVPR  = {CVPR})

@String(ICCV  = {ICCV})

@String(ECCV  = {ECCV})

@String(NIPS  = {NeurIPS})

@String(ICLR  = {ICLR})

@String(ICML  = {ICML})

@String(PAMI = {IEEE Transactions Pattern Analysis Machine Intelligence})

@String(CVPR= {IEEE Conference on Computer Vision and Pattern Recognition})

@String(ICCV= {Int. Conference on Computer Vision})

@String(ECCV= {European Conference on Computer Vision})

@String(NIPS= {Advances in Neural Information Processing Systems})

@String(ICLR = {Int. Conference on Learning Representations})

@String(ICML = {Int. Conference on Machine Learning})

@inproceedings{lfcbm,
author={T. Oikarinen and S. Das and L. M. Nguyen and T. W. Weng},
title = {Label-Free Concept Bottleneck Models},
booktitle=ICLR,
year = 2023}

@inproceedings{he2016resnet,
author={K. He and X. Zhang and S. Ren and J. Sun},
title = {Deep residual learning for image recognition},
booktitle = CVPR,
year = 2016}

@inproceedings{srivastava2024vlgcbm,
title={Vlg-cbm: Training concept bottleneck models with vision-language guidance},
author={D. Srivastava and G. Yan and L. Weng},
booktitle=NIPS,
year=2024
}

@inproceedings{koh2020cbm,
author={P. W. Koh and T. Nguyen and Y. S. Tang and S. Mussmann and E. Pierson and B. Kim and P. Liang},
title = {Concept Bottleneck Models},
booktitle = ICML,
year = 2020}

@inproceedings{dncbm,
author={S. Rao and S. Mahajan and M. Böhle and B. Schiele},
title = {Discover-then-Name: Task-Agnostic Concept Bottlenecks via Automated Concept Discovery},
booktitle = ECCV,
year = 2024}

@inproceedings{IN,
author={J. Deng and W. Dong and R. Socher and L. J. Li and K. Li and L. Fei-Fei},
title = {Imagenet: A large-scale hierarchical image database},
booktitle = CVPR,
year = 2009}

@inproceedings{GPT3,
author={T. B. Brown and B. Mann and N. Ryder and M. Subbiah and J. Kaplan and P. Dhariwal and \etal},
title = {Language models are few-shot learners},
booktitle = NIPS,
year = 2020}

@inproceedings{post_hoc_cbm,
author={M. Yuksekgonul and M. Wang and J. Zou},
title = {Post-hoc concept bottleneck models},
booktitle = ICLR,
year = 2023}

@inproceedings{labo,
author={Y. Yang and A. Panagopoulou and S. Zhou and D. Jin and C. Callison-Burch and M. Yatskar},
title = {Language in a Bottle: Language Model Guided Concept Bottlenecks for Interpretable Image Classification},
booktitle = CVPR,
year = 2023}

@inproceedings{caron2021dinov2,
author={M. Caron and H. Touvron and I. Misra and H. Jegou and J. Mairal and P. Bojanowski and A. Joulin},
title = {Emerging properties in self-supervised vision transformers},
booktitle = ICCV,
year = 2021}

@inproceedings{zhai2023siglip,
title={Sigmoid Loss for Language Image Pre-Training},
author={X. Zhai and B. Mustafa and A. Kolesnikov and L. Beyer},
booktitle=ICCV,
year=2023
}

@inproceedings{xiao2025flair,
title={FLAIR: VLM with Fine-grained Language-informed Image Representations},
author={R. Xiao and S. Kim and M. I. Georgescu and Z. Akata and S. Alaniz},
booktitle=CVPR,
year=2025
}

@inproceedings{zhang2025sail,
title={Assessing and Learning Alignment of Unimodal Vision and Language Models},
author={L. Zhang and Q. Yang and A. Agrawal},
booktitle=CVPR,
year=2025
}

@inproceedings{clip,
author={A. Radford and J. W. Kim and C. Hallacy and A. Ramesh and G. Goh and S. Agarwal and G. Sastry and A. Askell and P. Mishkin and J. Clark and G. Krueger and I. Sutskever},
title = {Learning transferable visual models from natural language supervision},
booktitle = ICML,
year = 2021}

@article{spacy,
author={M. Honnibal and I. Montani and S. Van Landeghem and A. Boyd},
title = {spaCy: Industrial-strength Natural Language Processing in Python},
doi= {10.5281/zenodo.1212303},
year = 2020}

@article{cifar,
author={A. Krizhevsky},
title = {Learning multiple layers of features from tiny images},
year = 2009}

@article{cub,
author={C. Wah and S. Branson and P. Welinder and P. Perona and S. Belongie},
title = {The caltech-ucsd birds-200-2011 dataset},
journal = {Technical Report CNS-TR-2011-001, California Institute of Technology},
year = 2011}

@article{places,
author={B. Zhou and A. Lapedriza and A. Khosla and A. Oliva and A. Torralba},
title = {Places: A 10 million image database for scene recognition},
journal= PAMI,
year = 2017}

@inproceedings{margeloiu2021leakage,
title={Do concept bottleneck models learn as intended?},
author={A. Margeloiu and M. Ashman and U. Bhatt and Y. Chen and M. Jamnik and A. Weller},
booktitle={Int. Conference on Learning Representations Workshop on Responsible AI},
year={2021}
}

@inproceedings{mahinpei2021leakage,
title={Promises and pitfalls of black-box concept learning models},
author={A. Mahinpei and J. Clark and I. Lage and F. Doshi-Velez and W. Pan},
booktitle={Int. Conference on Machine Learning Workshop on Theoretic Foundation, Criticism, and Application Trend of Explainable AI},
year={2021}
}

@inproceedings{havasi2022leakage,
title={Addressing Leakage in Concept Bottleneck Models},
author={M. Havasi and S. Parbhoo and F. Doshi-Velez},
booktitle=NIPS,
year={2022}
}

@article{sun2024leakage,
title={Eliminating Information Leakage in Hard Concept Bottleneck Models with Supervised, Hierarchical Concept Learning},
author={A. Sun and Y. Yuan and P. Ma and S. Wang},
journal={arXiv preprint arXiv:2404.05945},
year={2024}
}

@article{parisini2025leakage,
title={Leakage and Interpretability in Concept-Based Models},
author={E. Parisini and T. Chakraborti},
journal={arXiv preprint arXiv:2504.14094},
year={2025}
}

@article{makonnen2025measuring,
title={Measuring leakage in concept-based methods: An information theoretic approach},
author={M. Makonnen and M. Vandenhirtz and S. Laguna and J. E. Vogt},
journal={arXiv preprint arXiv:2504.09459},
year={2025}
}

@inproceedings{kim2018cav,
author={B. Kim and M. Wattenberg and J. Gilmer and C. Cai and J. Wexler and F. Viegas and \etal},
title = {Interpretability beyond feature attribution: Quantitative testing with concept activation vectors (tcav)},
booktitle = ICML,
year = 2018}

@article{debot2025tradeoff,
author = {D. Debot and G. Marra},
title = {Quantifying the Accuracy-Interpretability Trade-Off in Concept-Based Sidechannel Models},
journal = {arXiv preprint arXiv:2510.05670},
year = 2025
}

@inproceedings{yan2023lm4cv,
title={Learning concise and descriptive attributes for visual recognition},
author={A. Yan and Y. Wang and Y. Zhong and C. Dong and Z. He and Y. Lu and W. Y. Wang and J. Shang and J. McAuley},
booktitle=ICCV,
pages={3090--3100},
year={2023}
}

@article{liu2023grdino,
title={Grounding DINO: Marrying DINO with Grounded Pre-Training for Open-Set Object Detection},
author={S. Liu and Z. Zeng and T. Ren and F. Li and H. Zhang and J. Yang and Q. Jiang and C. Li and J. Yang and H. Su and J. Zhu and L. Zhang},
journal={arXiv preprint arXiv:2303.05499},
year={2023}
}

@article{bolya2025pe,
title={Perception Encoder: The Best Visual Embeddings are not at the Output of the Network},
author={D. Bolya and P. Y. Huang and P. Sun and J. H. Cho and A. Madotto and C. Wei and T. Ma and J. Zhi and J. Rajasegaran and H. Rasheed and \etal},
journal={arXiv preprint arXiv:2504.13181},
year=2025
}

@article{venkataramanan2025franca,
title={Franca: Nested Matryoshka Clustering for Scalable Visual Representation Learning},
author={S. Venkataramanan and V. Pariza and M. Salehi and L. Knobel and S. Gidaris and E. Ramzi and A. Bursuc and Y. M. Asano},
journal={arXiv preprint arXiv:2507.14137},
year={2025}
}

@article{Debole25,
author={N. Debole and P. Barbiero and F. Giannini and A. Passerini and S. Teso and E. Marconato},
title={If Concept Bottlenecks are the Question, are Foundation Models the Answer?},
journal={arXiv preprint arXiv:2504.19774},
year={2025}
}

@article{prasse2024dcbm,
title={DCBM: Data-Efficient Visual Concept Bottleneck Models},
author={K. Prasse and P. Knab and S. Marton and C. Bartelt and M. Keuper},
journal={arXiv preprint arXiv:2412.11576},
year={2024}
}

@inproceedings{StammerWSK24,
author={W. Stammer and A. Wüst and D. Steinmann and K. Kersting},
title={Neural Concept Binder},
booktitle=NIPS,
year={2024}
}

@article{poeta2023concept,
title={Concept-based explainable artificial intelligence: A survey},
author={E. Poeta and G. Ciravegna and E. Pastor and T. Cerquitelli and E. Baralis},
journal={ACM Computing Surveys},
year={2023},
publisher={ACM New York, NY}
}

@article{wong2021sparse,
title={Leveraging Sparse Linear Layers for Debuggable Deep Networks},
author={E. Wong and S. Santurkar and A. M{\k{a}}dry},
journal={arXiv preprint arXiv:2105.04857},
year={2021}
}

@inproceedings{StammerSK21,
author={W. Stammer and P. Schramowski and K. Kersting},
title={Right for the Right Concept: Revising Neuro-Symbolic Concepts by Interacting With Their Explanations},
booktitle=CVPR,
pages={3619--3629},
year={2021}
}
}

\clearpage
\setcounter{page}{1}
\appendix
\maketitlesupplementary

\section{Supplementary}


\subsection{Additional Experimental Setup Details}
\label{sup:configurations}

In \autoref{tab:hyperparameters}, we report the learning rates ($\eta$) used for optimizing the concept encoder and classifier, along with the distillation temperature and the $\alpha$/$\beta$ weights for balancing the classifier loss terms. The table also indicates whether a learning rate scheduler is used. When needed, we employ a cosine annealing schedule with a minimum learning rate of $0.0001$ and a cycle length of 20 epochs. Hyperparameters are provided per dataset, and the same settings are used across different vision backbones and VLMs.

The Non-linear CBM in \autoref{fig:linearity} and the Distilled CBM in \autoref{fig:distillation} correspond to the same underlying model. For the Linear CBM and Vanilla CBM ablation experiments, we use the same hyperparameters for consistency.

\begin{table}[]
\centering
\caption{Hyperparameter settings for each dataset, including learning rates ($\eta$) and scheduler usage for the concept encoder and classifier, along with the $\alpha$ and $\beta$ loss weights and distillation temperature Identical settings are used across all vision backbones and VLMs.}
\label{tab:hyperparameters}
\vspace{0.5em}
\resizebox{0.98\linewidth}{!}{%
\begin{tabular}{@{}l | *{2}{c}@{} | *{5}{c}@{}}
\toprule
\textbf{Dataset} & \multicolumn{2}{c}{\textbf{Concept Encoder}} & \multicolumn{5}{c}{\textbf{Classifier}} \\
\cmidrule(lr){2-8}
& $\eta$ & Scheduler & $\alpha$ & $\beta$ & Temperature & $\eta$ & Scheduler\\
\midrule
ImageNet100   & 0.001 & \checkmark & 1 & 1 & 2 & 0.0001 & - \\
CUB200        & 0.0001 & - & 1 & 0.5 & 4 & 0.001 & \checkmark \\
Places365     & 0.0001 & - & 1 & 2 & 2 & 0.001 & \checkmark \\
CIFAR100      & 0.001 & \checkmark & 1 & 1 & 4 & 0.0001 & -\\
\bottomrule
\end{tabular}
}
\end{table}

\subsection{Irrelevant and Random Words Concept Sets}

We provide the randomly selected concepts from our irrelevant and random words concept sets in \autoref{tab:concept_sets}. These concept sets are used throughout all experiments presented in \autoref{fig:issues_lin}, \autoref{fig:concept_act_dist_sail}, \autoref{fig:linearity} and in \autoref{tab:goodness_concepts}. The Roman Law concepts were extracted from the corresponding Wikipedia page, while the random strings were sampled uniformly at random, as described in \autoref{experiments}.

\begin{table}[]
\centering
\caption{Sample of 30 concepts from the two concept sets used in our experiments. Roman Law concepts extracted from the content of the relevant Wikipedia page, and a set of randomly generated strings.}
\resizebox{\linewidth}{!}{
\begin{tabular}{l l | l l}
\multicolumn{2}{c}{\textbf{Roman Law Concepts}} & \multicolumn{2}{c}{\textbf{Random Words}}\\
 \midrule
Jurisprudential & historians & gsankm & sjklrveogjc \\
administrative & impede & ehxrpq & pdbhhidbnvjig \\
bureaucratization & individual & swsbkzguqjil & cygmao\\
centuries & iudicum & odbhhqr & foqmegkdcp \\
civil &  knowledge & rwkkmbezusn & fdqumo \\
delegation & latitude & qkfns & hkeeyhw \\
descendants & laws & ntqwbpkwald & kzhqpxhr\\
disappear &  legal & jnplmlek & sgwftaole \\
ecclesiastical & magistrate & qtxvhsjlex & esdyegcbw \\
emperors & mancipatio & qfggdblkkli & zghijqngltcoo \\
enjoy & military & nckusvorw & tihgu \\
faithful & more & znnufioys & dmcmmjvd \\
features & nations & kvqpasqqkgt & ghhtwblxlmphi \\
formularies & normal & ydatcmpeqsmnf & ulcgplindw \\
grants & obvious & zebwdzhjfx & cynrhdme\\
\vspace{1.5em}
\end{tabular} }
\label{tab:concept_sets}
\end{table}

\subsection{Concept Activation Distributions of Different VLMs}
\label{sup:vlm_dist}

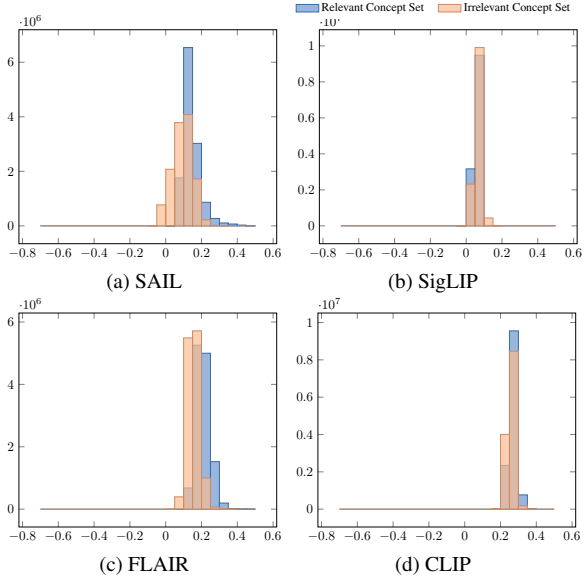
\begin{figure}
\begin{subfigure}[b]{.45\linewidth}
\centering
\begin{tikzpicture}[scale=0.5, every node/.append style={transform shape}]
    \begin{axis}[
        xtick={-0.4, -0.45,...,0.55},
        area style,
        ]
        \addplot+[draw=NavyBlue!80!black,
        fill=NavyBlue!60, fill opacity= 0.7, ybar interval,mark=no] plot coordinates { 
            (-0.7, 0)
            (-0.6499999999999999, 0)
            (-0.5999999999999999, 0)
            (-0.5499999999999998, 0)
            (-0.4999999999999998, 0)
            (-0.44999999999999973, 0)
            (-0.3999999999999997, 0)
            (-0.34999999999999964, 0)
            (-0.2999999999999996, 0)
            (-0.24999999999999956, 0)
            (-0.1999999999999995, 0)
            (-0.14999999999999947, 0)
            (-0.09999999999999942, 0)
            (-0.04999999999999938, 0)
            (6.661338147750939e-16, 153)
            (0.05000000000000071, 1759720)
            (0.10000000000000075, 6538073)
            (0.1500000000000008, 3025437)
            (0.20000000000000084, 865529)
            (0.2500000000000009, 272127)
            (0.30000000000000093, 107218)
            (0.350000000000001, 67862)
            (0.400000000000001, 28560)
            (0.45000000000000107, 3626)
            (0.5000000000000011, 563)
        };
        \addplot+[draw=Peach!80!black,
        fill=Peach!60, fill opacity= 0.7, ybar interval,mark=no] plot coordinates { 
            (-0.7, 0)
            (-0.6499999999999999, 0)
            (-0.5999999999999999, 0)
            (-0.5499999999999998, 0)
            (-0.4999999999999998, 0)
            (-0.44999999999999973, 0)
            (-0.3999999999999997, 0)
            (-0.34999999999999964, 0)
            (-0.2999999999999996, 0)
            (-0.24999999999999956, 0)
            (-0.1999999999999995, 0)
            (-0.14999999999999947, 0)
            (-0.09999999999999942, 30)
            (-0.04999999999999938, 770400)
            (6.661338147750939e-16, 2077293)
            (0.05000000000000071, 3785925)
            (0.10000000000000075, 4079251)
            (0.1500000000000008, 1722910)
            (0.20000000000000084, 223756)
            (0.2500000000000009, 9237)
            (0.30000000000000093, 98)
            (0.350000000000001, 0)
            (0.400000000000001, 0)
            (0.45000000000000107, 0)
            (0.5000000000000011, 0)
        };        

    \end{axis}
\end{tikzpicture} 
\subcaption{SAIL}
\end{subfigure}
\begin{subfigure}[b]{.45\linewidth}
\centering
\begin{tikzpicture}[scale=0.5, every node/.append style={transform shape}]
    \begin{axis}[
        xtick={-0.4, -0.45,...,0.55},
        area style,
        legend style={
        draw=none,
        at={(1,1.15)},
        legend columns=3,
        font=\small,
        /tikz/every even column/.append style={column sep=0.3cm}
        },
        ]
        \addplot+[draw=NavyBlue!80!black,
        fill=NavyBlue!60, fill opacity= 0.7, ybar interval,mark=no] plot coordinates {
            (-0.7, 0)
            (-0.6499999999999999, 0)
            (-0.5999999999999999, 0)
            (-0.5499999999999998, 0)
            (-0.4999999999999998, 0)
            (-0.44999999999999973, 0)
            (-0.3999999999999997, 0)
            (-0.34999999999999964, 0)
            (-0.2999999999999996, 0)
            (-0.24999999999999956, 0)
            (-0.1999999999999995, 0)
            (-0.14999999999999947, 0)
            (-0.09999999999999942, 0)
            (-0.04999999999999938, 334)
            (6.661338147750939e-16, 3166434)
            (0.05000000000000071, 9477038)
            (0.10000000000000075, 25092)
            (0.1500000000000008, 2)
            (0.20000000000000084, 0)
            (0.2500000000000009, 0)
            (0.30000000000000093, 0)
            (0.350000000000001, 0)
            (0.400000000000001, 0)
            (0.45000000000000107, 0)
            (0.5000000000000011, 0)
        };  
        \addplot+[draw=Peach!80!black,
        fill=Peach!60, fill opacity= 0.7, ybar interval,mark=no] plot coordinates {
            (-0.7, 0)
            (-0.6499999999999999, 0)
            (-0.5999999999999999, 0)
            (-0.5499999999999998, 0)
            (-0.4999999999999998, 0)
            (-0.44999999999999973, 0)
            (-0.3999999999999997, 0)
            (-0.34999999999999964, 0)
            (-0.2999999999999996, 0)
            (-0.24999999999999956, 0)
            (-0.1999999999999995, 0)
            (-0.14999999999999947, 0)
            (-0.09999999999999942, 0)
            (-0.04999999999999938, 1890)
            (6.661338147750939e-16, 2320556)
            (0.05000000000000071, 9907256)
            (0.10000000000000075, 438697)
            (0.1500000000000008, 501)
            (0.20000000000000084, 0)
            (0.2500000000000009, 0)
            (0.30000000000000093, 0)
            (0.350000000000001, 0)
            (0.400000000000001, 0)
            (0.45000000000000107, 0)
            (0.5000000000000011, 0)
        };
    \legend{Relevant Concept Set, Irrelevant Concept Set}
    \end{axis}
\end{tikzpicture}
\subcaption{SigLIP}
\end{subfigure}

\begin{subfigure}[b]{.45\linewidth}
\centering
\begin{tikzpicture}[scale=0.5, every node/.append style={transform shape}]
    \begin{axis}[
        xtick={-0.4, -0.45,...,0.55},
        area style,
        ]
        \addplot+[draw=NavyBlue!80!black,
        fill=NavyBlue!60, fill opacity= 0.7, ybar interval,mark=no] plot coordinates {
            (-0.7, 0)
            (-0.6499999999999999, 0)
            (-0.5999999999999999, 0)
            (-0.5499999999999998, 0)
            (-0.4999999999999998, 0)
            (-0.44999999999999973, 0)
            (-0.3999999999999997, 0)
            (-0.34999999999999964, 0)
            (-0.2999999999999996, 0)
            (-0.24999999999999956, 0)
            (-0.1999999999999995, 0)
            (-0.14999999999999947, 0)
            (-0.09999999999999942, 0)
            (-0.04999999999999938, 0)
            (6.661338147750939e-16, 0)
            (0.05000000000000071, 2804)
            (0.10000000000000075, 678397)
            (0.1500000000000008, 5255289)
            (0.20000000000000084, 4998419)
            (0.2500000000000009, 1523121)
            (0.30000000000000093, 196161)
            (0.350000000000001, 14111)
            (0.400000000000001, 589)
            (0.45000000000000107, 9)
            (0.5000000000000011, 0)
        };
        \addplot+[draw=Peach!80!black,
        fill=Peach!60, fill opacity= 0.7, ybar interval,mark=no] plot coordinates {
            (-0.7, 0)
            (-0.6499999999999999, 0)
            (-0.5999999999999999, 0)
            (-0.5499999999999998, 0)
            (-0.4999999999999998, 0)
            (-0.44999999999999973, 0)
            (-0.3999999999999997, 0)
            (-0.34999999999999964, 0)
            (-0.2999999999999996, 0)
            (-0.24999999999999956, 0)
            (-0.1999999999999995, 0)
            (-0.14999999999999947, 0)
            (-0.09999999999999942, 0)
            (-0.04999999999999938, 0)
            (6.661338147750939e-16, 390)
            (0.05000000000000071, 394996)
            (0.10000000000000075, 5490535)
            (0.1500000000000008, 5715766)
            (0.20000000000000084, 998907)
            (0.2500000000000009, 65643)
            (0.30000000000000093, 2593)
            (0.350000000000001, 70)
            (0.400000000000001, 0)
            (0.45000000000000107, 0)
            (0.5000000000000011, 0)
        };
    \end{axis}
\end{tikzpicture}
\subcaption{FLAIR}
\end{subfigure}
\begin{subfigure}[b]{.45\linewidth}
\centering
\begin{tikzpicture}[scale=0.5, every node/.append style={transform shape}]
    \begin{axis}[
        xtick={-0.4, -0.45,...,0.55},
        area style,
        ]
        \addplot+[draw=NavyBlue!80!black,
        fill=NavyBlue!60, fill opacity= 0.7, ybar interval,mark=no] plot coordinates {
            (-0.7, 0)
            (-0.6499999999999999, 0)
            (-0.5999999999999999, 0)
            (-0.5499999999999998, 0)
            (-0.4999999999999998, 0)
            (-0.44999999999999973, 0)
            (-0.3999999999999997, 0)
            (-0.34999999999999964, 0)
            (-0.2999999999999996, 0)
            (-0.24999999999999956, 0)
            (-0.1999999999999995, 0)
            (-0.14999999999999947, 0)
            (-0.09999999999999942, 0)
            (-0.04999999999999938, 0)
            (6.661338147750939e-16, 0)
            (0.05000000000000071, 0)
            (0.10000000000000075, 0)
            (0.1500000000000008, 2242)
            (0.20000000000000084, 2341710)
            (0.2500000000000009, 9553152)
            (0.30000000000000093, 765569)
            (0.350000000000001, 6227)
            (0.400000000000001, 0)
            (0.45000000000000107, 0)
            (0.5000000000000011, 0)
        };
        \addplot+[draw=Peach!80!black,
        fill=Peach!60, fill opacity= 0.7, ybar interval,mark=no] plot coordinates {
            (-0.7, 0)
            (-0.6499999999999999, 0)
            (-0.5999999999999999, 0)
            (-0.5499999999999998, 0)
            (-0.4999999999999998, 0)
            (-0.44999999999999973, 0)
            (-0.3999999999999997, 0)
            (-0.34999999999999964, 0)
            (-0.2999999999999996, 0)
            (-0.24999999999999956, 0)
            (-0.1999999999999995, 0)
            (-0.14999999999999947, 0)
            (-0.09999999999999942, 0)
            (-0.04999999999999938, 0)
            (6.661338147750939e-16, 0)
            (0.05000000000000071, 0)
            (0.10000000000000075, 0)
            (0.1500000000000008, 9995)
            (0.20000000000000084, 4007777)
            (0.2500000000000009, 8463080)
            (0.30000000000000093, 188024)
            (0.350000000000001, 24)
            (0.400000000000001, 0)
            (0.45000000000000107, 0)
            (0.5000000000000011, 0)
        };
    \end{axis}
\end{tikzpicture}
\subcaption{CLIP}
\end{subfigure}
\caption{Histogram of concept activations computed with different VLMs for the relevant and irrelevant concept sets on ImageNet100, highlighting differences in activation distributions across models when evaluated on meaningful versus random concepts.}
\label{fig:concept_act_dist}
\end{figure}

We compute concept activations on the ImageNet100 dataset using both the GPT-generated concept set as the relevant concepts and a set of random strings as the irrelevant concepts. The evaluated VLMs include CLIP, SigLIP, SAIL, and FLAIR. For each model, we report the distribution of the top-100 concept activations per image in \autoref{fig:concept_act_dist}. Across the four VLMs, the activation distributions differ substantially in both scale and spread. SAIL exhibits the widest distribution, with both relevant and irrelevant concepts producing high activations, whereas SigLIP and CLIP show much more compact distributions. Overall, these results highlight the differences in how VLMs encode concepts.

\subsection{Goodness of Concepts with Varying Cutoff Values}
\label{sup:cutoff_values}
We use a cutoff of 100 concepts that can be consistently applied to both ImageNet100 and CUB200, whose concept sets contain 4,751 and 370 concepts, respectively. Importantly, the goodness-of-concepts metric is largely insensitive to the specific choice of cutoff: relevant concept sets consistently exhibit lower entropy, whereas irrelevant concept sets yield higher entropy across different cutoff values.
In \autoref{tab:cutoff}, we provide a quantitative evaluation of goodness of concepts on ImageNet100 with varying cutoff values. We observe that varying cutoff values yields robust behavior.

\begin{table}[]
\centering
\caption{Robustness of the goodness of concepts metric to varying cutoff values on ImageNet100.}
\label{tab:cutoff}
\begin{tabular}{l | c c c }
    \multicolumn{4}{c}{\textbf{Task Agnostic} $\downarrow$}\\
    \textbf{Concept Set}   & \textbf{Top-10} & \textbf{Top-100} & \textbf{Top-250}  \\
    \hline
    Relevant             & \textbf{1.751} & \textbf{3.660} & \textbf{4.521}  \\
    Irrelevant           & 2.254 & 4.504 & 5.376  \\
    Random               & 2.284 & 4.565 & 5.465   \\
\end{tabular}
\end{table}

\subsection{Goodness of Concepts Results with Various VLMs}
\label{sup:GoC}

The goodness of concept metric is calculated across ImageNet100, CUB200 and Places365 datasets using several VLMs. Results are reported in two settings: task-agnostic in \autoref{tab:goodness_taskagn} and task-specific in \autoref{tab:goodness_taskspe}.
 
Across all datasets and models, the metric consistently demonstrate a clear separation between relevant and irrelevant concepts. Relevant concepts achieve the lowest entropy, indicating stronger alignment between the concepts and the images in both settings. 

This trend is especially pronounced for SAIL, which also exhibits a distinctive concept activation distribution in \autoref{fig:concept_act_dist}. SAIL yields markedly lower entropy values for relevant concepts in both ImageNet100 and Places365. 

Overall, these results confirm that the goodness of concepts metric reliably captures concept set suitability across a range of VLMs and diverse domains.

\begin{table}[]
\centering
\caption{Task-agnostic goodness of concepts: suitability of different concept sets for different datasets. Relevant concepts are GPT-generated in ImageNet100 and Places365 and ground-truth attributes in CUB200. The irrelevant concept set consists of Roman Law terminology. Random strings denote non-word concepts. Lower entropy indicates better alignment.}
\resizebox{\linewidth}{!}{
\begin{tabular}{l|cccc}
\multicolumn{5}{c}{\textbf{ImageNet100}}  \\

\textbf{Concept Set} & CLIP & SAIL & FLAIR & SigLIP  \\
 \midrule
Relevant             & 4.479  & 3.660 & 4.545 & 4.494  \\
Irrelevant           & 4.480 & 4.504 & 4.511 & 4.534  \\
Random words         & 4.555 & 4.565 & 4.565 & 4.522 
\vspace{1.5em}
\end{tabular}  }
\resizebox{\linewidth}{!}{
\begin{tabular}{l|cccc}
\multicolumn{5}{c}{\textbf{CUB200}} \\

\textbf{Concept Set} & CLIP & SAIL & FLAIR & SigLIP  \\
 \midrule
Relevant            & 4.456 & 4.305 & 4.459 & 4.463 \\
Irrelevant          & 4.422 & 4.491 & 4.533 & 4.534  \\
Random words        & 4.542 & 4.553 & 4.577 & 4.491 
\vspace{1.5em}
\end{tabular}  }
\resizebox{\linewidth}{!}{
\begin{tabular}{l|cccc}
\multicolumn{5}{c}{\textbf{Places365}} \\

\textbf{Concept Set} & CLIP & SAIL & FLAIR & SigLIP  \\
 \midrule
Relevant             & 4.482 & 3.077 & 4.534 & 4.512 \\
Irrelevant           & 4.472 & 4.505 & 4.531 & 4.539 \\
Random words         & 4.557 & 4.566 & 4.563 & 4.600

\end{tabular}   }
\label{tab:goodness_taskagn}
\end{table}

\begin{table}[]
\centering
\caption{Task-specific goodness of concepts: suitability of different concept sets for different datasets. Relevant concepts are GPT-generated in ImageNet100 and Places365 and ground-truth attributes in CUB200. The irrelevant concept set consists of Roman Law terminology. Random strings denote non-word concepts. Lower entropy indicates better alignment.}
\resizebox{\linewidth}{!}{
\begin{tabular}{l|cccc}
 \multicolumn{5}{c}{\textbf{ImageNet100}} \\
\textbf{Concept Set} & CLIP & SAIL & FLAIR & SigLIP \\
 \midrule
Relevant             & 4.517 & 3.738 & 4.578 & 4.522\\
Irrelevant           & 4.529 & 4.545 & 4.572 & 4.559 \\
Random words         & 4.576 & 4.579 & 4.589 & 4.556 
\vspace{1.5em}
\end{tabular}   }

\resizebox{\linewidth}{!}{
\begin{tabular}{l|cccc}
\multicolumn{5}{c}{\textbf{CUB200}} \\

\textbf{Concept Set} & CLIP & SAIL & FLAIR & SigLIP  \\
 \midrule
Relevant             & 4.456 & 4.383 & 4.459 & 4.463 \\
Irrelevant           & 4.422 & 4.526 & 4.533 & 4.534  \\
Random words         & 4.542 & 4.565 & 4.577 & 4.491 
\vspace{1.5em}
\end{tabular}   }

\resizebox{\linewidth}{!}{
\begin{tabular}{l|cccc}
\multicolumn{5}{c}{\textbf{Places365}} \\

\textbf{Concept Set} & CLIP & SAIL & FLAIR & SigLIP  \\
 \midrule
Relevant             & 4.549 & 2.889 & 4.579 & 4.512 \\
Irrelevant           & 4.551 & 4.547 & 4.583 & 4.540 \\
Random words         & 4.591 & 4.578 & 4.592 & 4.600

\end{tabular}   }
\label{tab:goodness_taskspe}
\end{table}

\subsection{Refining concept set via Goodness of Concepts } 
\label{sup:refining}
\begin{figure}[t]
\centering

\begin{tikzpicture}
  \begin{axis}[
    height=4cm,
    width=0.45\textwidth,
    xmin = 0, xmax = 70,
    ymin=5, ymax=5.45,
    axis y line* = left, 
    xlabel={\# of Removed Concepts},
    xlabel near ticks,
    xlabel style={font=\small\bfseries},
    ylabel={Entropy},
    ylabel near ticks,
    ylabel style={font=\small\bfseries},
    legend style={
            draw=none,
            font=\footnotesize,
            at={(0.5,-.35)},
            anchor=north,
            legend columns=3,
        },
    ]
    \addplot[
    color=BrickRed,
    line width=1.2pt,
    ]
    coordinates {
        (0,5.405) (10,5.369) (20,5.339) (30,5.307) 
        (40,5.266) (50,5.231) (60,5.188) (70,5.144)
    };
    \addplot[
    color=CadetBlue,
    line width=1.2pt,
    ]
    coordinates {
        (0,5.405) (10,5.360) (20,5.313) (30,5.264) 
        (40,5.213) (50,5.159) (60,5.103) (70,5.044)
    };
    \legend{Random-Removal, Entropy-Guided} 
  \end{axis}
  \pgfplotsset{every axis y label/.append style={rotate=180,yshift=9.5cm}}
  \begin{axis}[
        height=4cm,
        width=0.45\textwidth,
        xmin=0, xmax=70,
        ymin=0, ymax=25,
        hide x axis,
        axis y line*=right,
        ylabel={\# Removed Rand.-Words},
        ylabel near ticks,
        ylabel style={font=\small\bfseries},
    ]
    \addplot[
    color=BrickRed,
    line width=1.2pt,
    dashed,
    ]
    coordinates {
        (0,0) (10,1.9) (20,4.1) (30,6.4) 
        (40,9.2) (50,11.7) (60,14.8) (70,17.6)
    };
    \addplot[
    color=CadetBlue,
    line width=1.2pt,
    dashed,
    ]
    coordinates {
        (0,0) (10,4) (20,9) (30,10) 
        (40,11) (50,15) (60,17) (70,20)
    };
  \end{axis}
\end{tikzpicture}
\caption{Concept set refinement. Entropy-guided removal (blue) outperforms random removal (red, averaged over 10 trials) and preferentially eliminates random concepts, demonstrating the effectiveness of the goodness-of-concepts metric. Solid lines correspond to the entropy (left y-axis), while dashed lines correspond to the number of removed random words (right y-axis).}
\label{fig:refinement}
\end{figure}
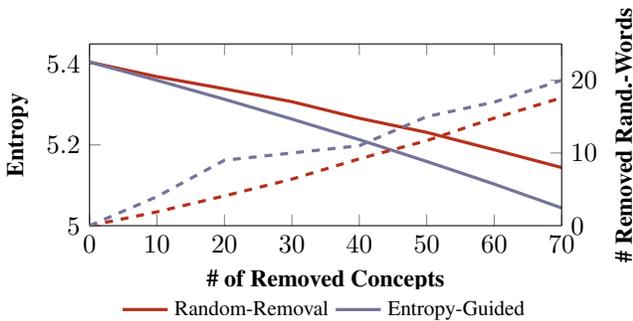

In \autoref{fig:refinement}, we demonstrate that the goodness of concepts metric enables concept set refinement. We create a mixed concept set with 75\% CUB attributes and 25\% random words.
The figure above shows entropy reduction over removal iterations. The red line represents random concept removal (averaged over 10 trials), while the blue line shows entropy-guided removal (selecting concepts whose removal minimizes entropy). Entropy-guided removal achieves consistently lower entropy, indicating more concentrated and meaningful concept activations.
Crucially, entropy-guided removal preferentially eliminates random words: 17.9\% of removed concepts came from the random-word subset, compared to only 13.7\% for random removal. This demonstrates that goodness of concepts effectively identifies and removes irrelevant concepts, validating its utility for concept set refinement.

\subsection{Ablations}
To inspect the disentangled effects of non-linearity and distillation, we conduct an ablation experiment. As shown in \autoref{tab:ablation},  linearity is the primary factor affecting performance. However, non-linearity is required for a trustworthy CBM (\cf \autoref{fig:linearity}). To reconcile this trade-off, we improve the accuracy of non-linear models through knowledge distillation, which consistently recovers a substantial portion of the performance gap.

\begin{table}[]
\centering
\caption{Ablation study disentangling the effects of linearity and knowledge distillation.}
\label{tab:ablation}
\begin{tabular}{l c}
Configuration & Accuracy (\%) \\
\hline
Linear & 87.96 \\
Linear + Distillation & 90.37 \\
Non-linear & 82.27 \\
Non-linear + Distillation & 85.56
\end{tabular}
\end{table}

\subsection{Extended Interpretability Results on Different Datasets}
\label{sup:qualitative}

\begin{figure*}[t!]
\centering
\begin{tikzpicture}
  \matrix[matrix of nodes, nodes={anchor=center}, column sep=0.5mm, row sep=1mm] {
     & \parbox{0.1\linewidth}{\centering\small Class Name} 
     & \parbox{0.1\linewidth}{\centering\small Top-5 Concepts} 
     & \parbox{0.1\linewidth}{\centering\small Class Name} 
     & \parbox{0.1\linewidth}{\centering\small Top-5 Concepts}
     & \parbox{0.1\linewidth}{\centering\small Class Name} 
     & \parbox{0.1\linewidth}{\centering\small Top-5 Concepts} 
     & \parbox{0.1\linewidth}{\centering\small Class Name} 
     & \parbox{0.1\linewidth}{\centering\small Top-5 Concepts}\\
     
    \parbox[t][0.15\linewidth]{0.01\linewidth}{\rotatebox{90}{\centering\small ImageNet100}} 
    & \parbox[t][0.15\linewidth]{0.1\linewidth}{\centering\tiny Robin \\ \includegraphics[height=0.8\linewidth]{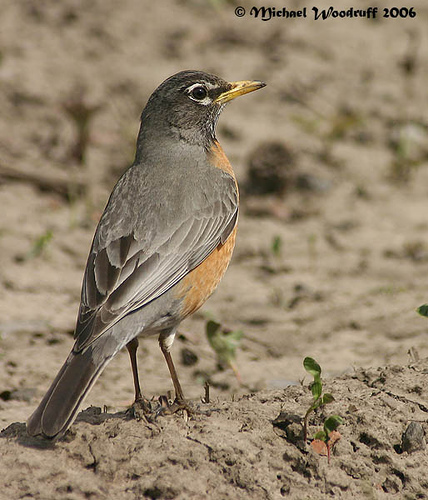} }
    & \parbox[t][0.15\linewidth]{0.11\linewidth}{\tiny 1. a bright orange breast \\ 2. a rufous back and wings \\ 3. a reddish-brown breast \\ 4. a small songbird \\ 5. finch} 
    &  \parbox[t][0.15\linewidth]{0.1\linewidth}{\centering\tiny Carbonara \\ \includegraphics[height=0.8\linewidth]{qualitative/carbonara.JPEG}} 
    & \parbox[t][0.15\linewidth]{0.11\linewidth}{\tiny 1. pasta \\ 2. bacon or pancetta \\ 3. Italian dish \\ 4. noodles \\ 5. a creamy sauce} 
    &  \parbox[t][0.15\linewidth]{0.1\linewidth}{\centering\tiny Pedestal \\ \includegraphics[height=0.8\linewidth]{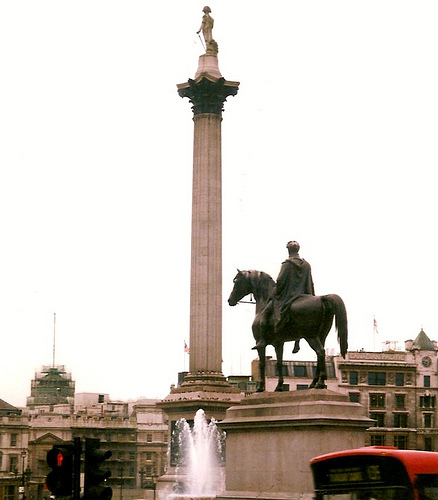}}
    & \parbox[t][0.15\linewidth]{0.11\linewidth}{\tiny 1. a statue \\ 2. a tall, upright stature \\ 3. a small stature \\ 4. columns or towers \\ 5. marble or stone construction} 
    &  \parbox[t][0.15\linewidth]{0.1\linewidth}{\centering\tiny Slide Rule \\ \includegraphics[height=0.8\linewidth]{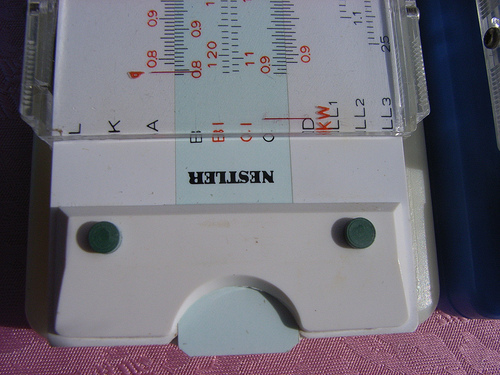}} 
    & \parbox[t][0.15\linewidth]{0.11\linewidth}{\tiny 1. rulers \\ 2. a foot measuring device \\ 3. a ruler \\ 4. a height chart \\ 5. a train schedule } \\
    
    \parbox[t][0.15\linewidth]{0.01\linewidth}{\rotatebox{90}{\centering\small CUB200}} 
    & \parbox[t][0.15\linewidth]{0.1\linewidth}{\centering\tiny Bay breasted Warbler \\ \includegraphics[height=0.7\linewidth]{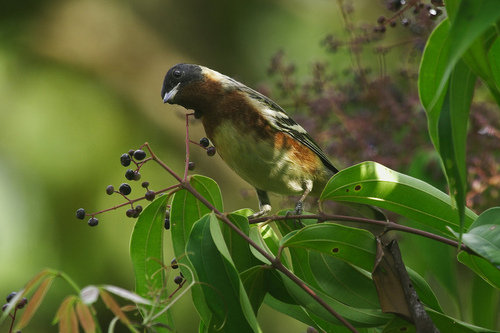} }
    & \parbox[t][0.15\linewidth]{0.11\linewidth}{\tiny 1. a long, tapered tail \\\ 2. a rust-colored cap and nape \\ 3. a pinkish breast \\ 4. a buffy breast \\ 5. a red spot on the beak} 
    &  \parbox[t][0.15\linewidth]{0.1\linewidth}{\centering\tiny Cactus Wren \\ \includegraphics[height=0.8\linewidth]{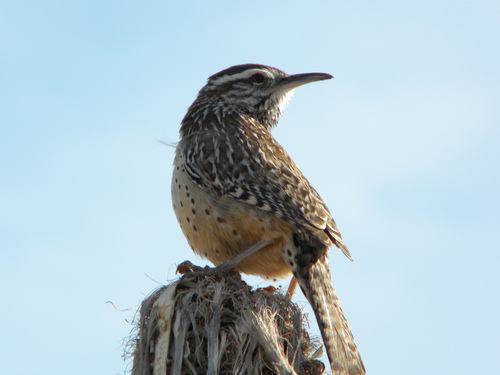}} 
    & \parbox[t][0.15\linewidth]{0.11\linewidth}{\tiny 1. long hind legs for jumping \\ 2. a white stripe on the wings \\ 3. mottled grey and white plumage \\ 4. dark wingtips \\ 5. a large, green head} 
    &  \parbox[t][0.15\linewidth]{0.1\linewidth}{\centering\tiny Purple Finch \\ \includegraphics[height=0.8\linewidth]{qualitative/Purple_Finch.jpg}}
    & \parbox[t][0.15\linewidth]{0.11\linewidth}{\tiny 1. a red head \\ 2. a rosy breast \\ 3. a pinkish breast \\ 4. a purple-red head and breast \\ 5. a white eye-ring} 
    &  \parbox[t][0.15\linewidth]{0.1\linewidth}{\centering\tiny Blue winged Warbler \\ \includegraphics[height=0.7\linewidth]{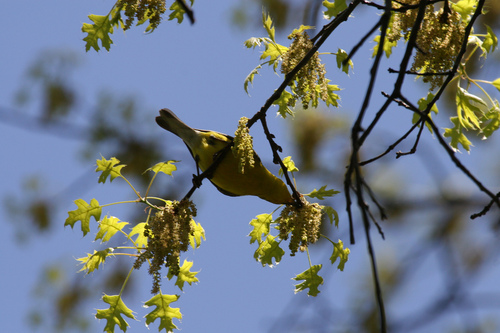}} 
    & \parbox[t][0.15\linewidth]{0.11\linewidth}{\tiny 1. a yellow head \\ 2. a yellow body \\ 3. a yellow head and throat \\ 4. a yellow cap on the head \\ 5. a brown back and wings} \\

    \parbox[t][0.15\linewidth]{0.01\linewidth}{\rotatebox{90}{\centering\small CIFAR100}} 
    & \parbox[t][0.15\linewidth]{0.1\linewidth}{\centering\tiny Road \\ \includegraphics[height=0.7\linewidth]{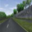} }
    & \parbox[t][0.15\linewidth]{0.11\linewidth}{\tiny 1. a long, straight path \\ 2. a driveway \\ 3. way \\ 4. a path \\ 5. markings for lanes} 
    &  \parbox[t][0.15\linewidth]{0.1\linewidth}{\centering\tiny Cloud \\ \includegraphics[height=0.8\linewidth]{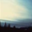}} 
    & \parbox[t][0.15\linewidth]{0.11\linewidth}{\tiny 1. the sky \\ 2. floating in the sky \\ 3. sky \\ 4. lots of smoke \\ 5. weather} 
    &  \parbox[t][0.15\linewidth]{0.1\linewidth}{\centering\tiny Telephone \\ \includegraphics[height=0.8\linewidth]{qualitative/telephone.png}}
    & \parbox[t][0.15\linewidth]{0.11\linewidth}{\tiny 1. a phone \\ 2. communication device \\ 3. a keypad on one side \\ 4. a phone book \\ 5. lateral symmetry} 
    &  \parbox[t][0.15\linewidth]{0.1\linewidth}{\centering\tiny Sweet Pepper \\ \includegraphics[height=0.7\linewidth]{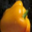}} 
    & \parbox[t][0.15\linewidth]{0.11\linewidth}{\tiny 1. vegetable \\ 2. a kitchen \\ 3. a compost bin \\ 4. a red, green, or yellow color \\ 5. a greenhouse} \\

    \parbox[t][0.15\linewidth]{0.01\linewidth}{\rotatebox{90}{\centering\small Places365}} 
    & \parbox[t][0.15\linewidth]{0.1\linewidth}{\centering\tiny Ocean \\ \includegraphics[height=0.7\linewidth]{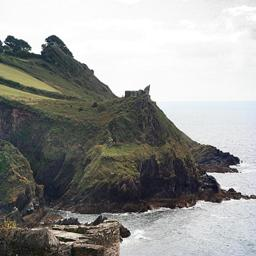} }
    & \parbox[t][0.15\linewidth]{0.11\linewidth}{\tiny 1. the sea \\ 2. clear blue waters \\ 3. seaweed \\ 4. a surfboard \\ 5. a boat ramp} 
    &  \parbox[t][0.15\linewidth]{0.1\linewidth}{\centering\tiny Auto Showroom \\ \includegraphics[height=0.8\linewidth]{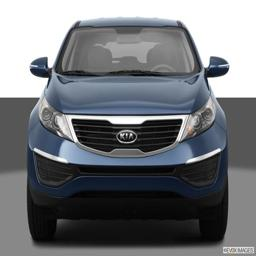}} 
    & \parbox[t][0.15\linewidth]{0.11\linewidth}{\tiny 1. car models \\ 2. cars parked in the lot \\ 3. vertical or nearly vertical \\ 4. a showroom floor \\ 5. a glossy finish} 
    &  \parbox[t][0.15\linewidth]{0.1\linewidth}{\centering\tiny Lake, natural \\ \includegraphics[height=0.8\linewidth]{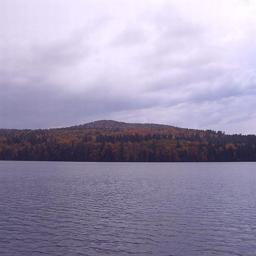}}
    & \parbox[t][0.15\linewidth]{0.11\linewidth}{\tiny 1. a body of water \\ 2. a reservoir \\ 3. a lake or river \\ 4. a wetland \\ 5. a koi pond} 
    &  \parbox[t][0.15\linewidth]{0.1\linewidth}{\centering\tiny Palace \\ \includegraphics[height=0.7\linewidth]{qualitative/palace.png}} 
    & \parbox[t][0.15\linewidth]{0.11\linewidth}{\tiny 1. a royal residence \\ 2. grand, ornate architecture \\ 3. often has towers or turrets \\ 4. guards \\ 5. a garden on top of a building } \\
  };
\end{tikzpicture}
\caption{Illustration of the Top-5 concepts, ranked by their contribution to the final prediction for randomly selected classes, along with representative images from ImageNet100, CIFAR100, CUB200, and Places365, highlighting model interpretability across diverse domains.}
\label{fig:top5concepts_extended}
\end{figure*}

The extended version of \autoref{fig:top5concepts} is provided in the \autoref{fig:top5concepts_extended} where we report the top-5 activated concepts for random classes. The best performing vision backbone and VLM combinations for each dataset are as follows: Perception Encoder + SigLIP for ImageNet100, Perception Encoder + CLIP for all remaining datasets.

\end{document}